\documentclass[11pt, a4paper, logo, copyright, nonumbering]{main}
\usepackage[authoryear, sort&compress, round]{natbib}
\usepackage{dblfloatfix}
\usepackage{ulem}
\usepackage{caption}
\usepackage{dramatist}
\usepackage{xspace}
\usepackage{pifont} %
\usepackage{multirow}
\usepackage{tcolorbox}
\usepackage{xltabular}
\usepackage{longtable}
\usepackage{standalone}
\usepackage{hyperref}
\usepackage{makecell}
\usepackage[ruled,vlined,linesnumbered]{algorithm2e}
\usepackage{amsfonts}
\usepackage{amsmath}
\usepackage{amssymb}
\usepackage{lineno}
\usepackage{multirow}
\usepackage{adjustbox}

\usepackage[bottom]{footmisc}

\usepackage{CJKutf8}
\usepackage{subfigure}
\usepackage{setspace}

\usepackage{dsfont}
\usepackage{array} %
\usepackage{tabularx} %
\usepackage{subfigure} %
\usepackage{xcolor} %
\usepackage{tabularx}
\usepackage{booktabs}
\usepackage{xspace}

\usepackage{lipsum}  %
\usepackage{multicol} %

\usepackage{cleveref}
\usepackage{epigraph}

\makeatletter
\def\@BTrule[#1]{%
  \ifx\longtable\undefined
    \let\@BTswitch\@BTnormal
  \else\ifx\hline\LT@hline
    \nobreak
    \let\@BTswitch\@BLTrule
  \else
    \let\@BTswitch\@BTnormal
  \fi\fi
  \global\@thisrulewidth=#1\relax
  \ifnum\@thisruleclass=\tw@\vskip\@aboverulesep\else
  \ifnum\@lastruleclass=\z@\vskip\@aboverulesep\else
  \ifnum\@lastruleclass=\@ne\vskip\doublerulesep\fi\fi\fi
  \@BTswitch
}
\makeatother

\addto\extrasenglish{
}

{\begin{list}{}%
        {\setlength{\leftmargin}{#1}}%
        \item[]%
}
{\end{list}}

\reportnumber{001} %

\title{\centering Step-GUI Technical Report }

\def\huggingface{\raisebox{-1.5pt}{\includegraphics[height=1.05em]{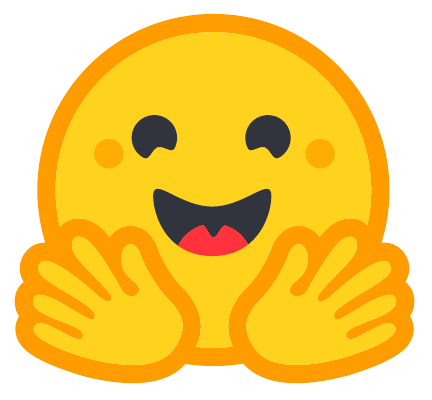}}}
\def\github{\raisebox{-1.5pt}{\includegraphics[height=1.05em]{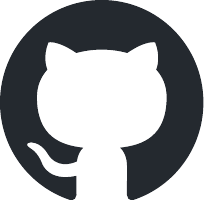}}}
\def\homepage{\raisebox{-1.5pt}{\includegraphics[height=1.2em]{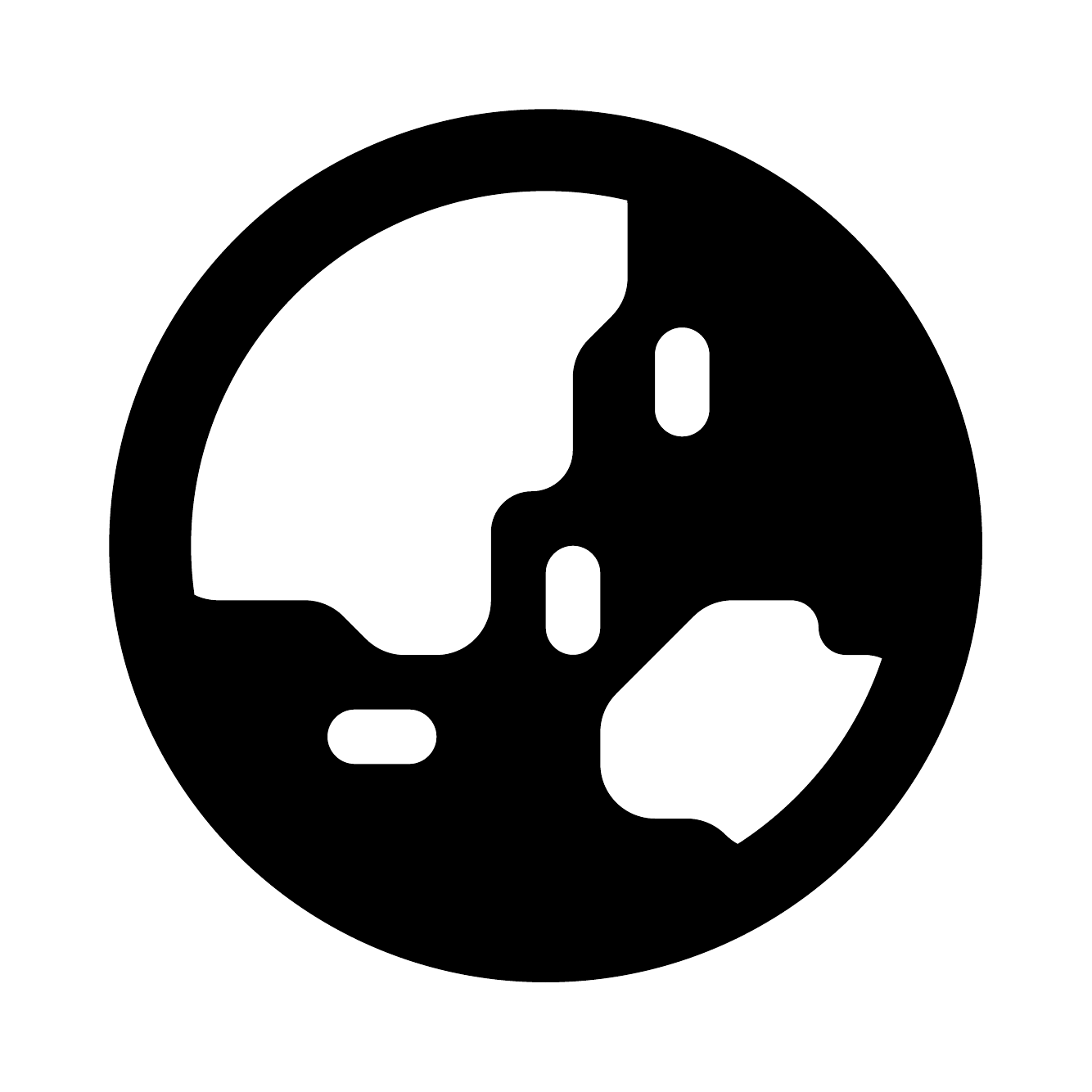}}}

\author[*]{
\textbf{GELab-Team, StepFun}
\\
\vspace{-0.7em}
\small \homepage~\textbf{Homepage}: \url{https://opengelab.github.io/} \\
\small \github~\textbf{Github}: \url{https://github.com/stepfun-ai/gelab-zero} \\
\small \huggingface~\textbf{Huggingface}: \href{https://huggingface.co/stepfun-ai/GELab-Zero-4B-preview}{Step-GUI Collections}
}

\renewcommand{\phi}{\varphi}

\renewcommand{\epsilon}{\varepsilon}
\renewcommand{\imath}{\mathrm{i}}

\newlength{\restsubwidth}
\newlength{\restsubheight}
\newlength{\restsubmoreheight}
\newcommand{\rest}[2]{%
        \settowidth{\restsubwidth}{\ensuremath{#2}}
        \settoheight{\restsubheight}{\ensuremath{{}_{#2}}}
        \ensuremath{{#1\hskip 0.5pt}_{\vrule\kern2pt\parbox[b][%
        4pt][b]{\the\restsubwidth}{%
                        \ensuremath{{}_{#2}}}}}
        }

\begin{abstract}
Recent advances in multimodal large language models unlock unprecedented opportunities for GUI automation. 
However, a fundamental challenge remains: how to efficiently acquire high-quality training data while maintaining annotation reliability? 
We introduce a self-evolving training pipeline powered by the \textbf{Calibrated Step Reward System}, which converts model-generated trajectories into reliable training signals through trajectory-level calibration, achieving $>$90\% annotation accuracy with 10\textasciitilde100$\times$ lower cost. 
Leveraging this pipeline, we introduce Step-GUI, a family of models (4B/8B) that achieves state-of-the-art GUI performance (8B: 80.2\% AndroidWorld, 48.5\% OSWorld, 62.6\% ScreenShot-Pro) while maintaining robust general capabilities.
As GUI agent capabilities improve, practical deployment demands standardized interfaces across heterogeneous devices while protecting user privacy. 
To this end, we propose \textbf{GUI-MCP}, the first Model Context Protocol for GUI automation with hierarchical architecture that combines low-level atomic operations and high-level task delegation to local specialist models, enabling high-privacy execution where sensitive data stays on-device. 
Finally, to assess whether agents can handle authentic everyday usage, we introduce \textbf{AndroidDaily}, a benchmark grounded in real-world mobile usage patterns with 3146 static actions and 235 end-to-end tasks across high-frequency daily scenarios (8B: static 89.91\%, end-to-end 52.50\%). 
Our work advances the development of practical GUI agents and demonstrates strong potential for real-world deployment in everyday digital interactions.

\end{abstract}

\begin{document}

\maketitle


\definecolor{colorfirst}{RGB}{252,141,89}
\definecolor{colorsecond}{RGB}{253,187,132}
\definecolor{colorthird}{RGB}{253,212,158}
\definecolor{colorfourth}{RGB}{254,232,200}
\definecolor{colorfifth}{RGB}{255,247,236}
\definecolor{myred}{RGB}{242,128,128}
\definecolor{mygreen}{RGB}{112,180,143}
\definecolor{myblue}{RGB}{210,225,255}
\definecolor{citypink}{RGB}{227,108,194}
\definecolor{cityblue}{RGB}{128,159,225}
\newcommand{\rankfirst}[0]{\cellcolor{colorfirst}}
\newcommand{\ranksecond}[0]{\cellcolor{colorsecond}}
\newcommand{\rankthird}[0]{\cellcolor{colorthird}}
\newcommand{\rankfourth}[0]{\cellcolor{colorfourth}}
\newcommand{\rankfifth}[0]{\cellcolor{colorfifth}}
\DeclareRobustCommand{\legendsquare}[1]{%
  \textcolor{#1}{\rule{2ex}{2ex}}%
}
\DeclareRobustCommand{\legendsquarebox}[1]{%
  \tikz[] \draw[black, fill=#1, line width=0.4pt] (0,0) rectangle (1.5ex,1.5ex);%
}
\newcommand{\cmark}{\textcolor{mygreen}{\ding{51}}}%
\newcommand{\xmark}{\textcolor{myred}{\ding{55}}}%


\begin{figure}[h]
    \centering
    \captionsetup{justification=justified, singlelinecheck=false}
    \includegraphics[
    width=1.0\linewidth
    ]{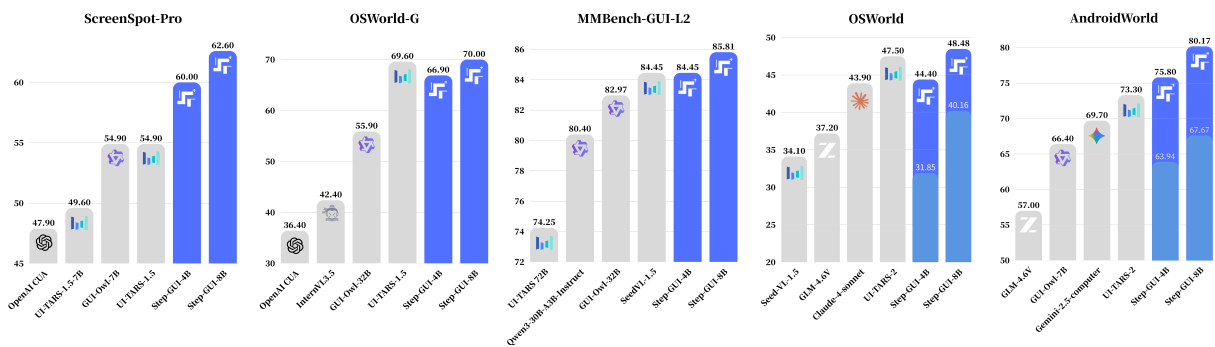}
    \caption{Performance Overview across Heterogeneous GUI Benchmarks. We compare Step-GUI (4B/8B) against leading baselines on five diverse benchmarks covering grounding (ScreenSpot-Pro, OSWorld-G, MMBench-GUI-L2) and end-to-end agentic tasks (OSWorld, AndroidWorld). End-to-end results use pass@3 metric to mitigate non-model-related failures (e.g., CAPTCHA, VM crashes). Pass@1 results are also shown in a different shade of blue.
    The results demonstrate that Step-GUI-8B achieves state-of-the-art performance, outperforming existing open-source and proprietary agents, even surpassing models with much larger parameter counts.
    }
    \label{fig:intro_head}
\end{figure}

\section{Introduction}

The advent of Large Language Models (LLMs) unlocks unprecedented capabilities in natural language understanding and generation, fundamentally transforming how machines process and generate human language. Recent breakthroughs in multimodal large language models~\cite{Bai2025Qwen3VL, comanici2025gemini_gemini25, gpt5, claudeopus45, guo2025seed1_seed15} further extend these capabilities to visual perception and action planning, providing the foundational abilities necessary for GUI automation. These advances enable models to understand complex visual interfaces and reason about appropriate interactions, opening new possibilities for autonomous GUI agents that can assist users in their daily digital tasks.

Building on these foundational capabilities, the research community makes significant progress in GUI agent development. Various approaches explore methods for acquiring GUI interaction data~\cite{gao2024mobileviews, rawles2023androidinthewild, liu2025scalecua, wang2025opencua}, establishing critical data infrastructure for training GUI agents. Several powerful GUI agent models~\cite{qin2025ui_uitars1, wang2025ui_uitars2, ye2025mobile_mobileagentv3, zeng2025uitron} demonstrate impressive capabilities in navigating and manipulating graphical interfaces. However, a critical challenge remains: \textit{how to efficiently obtain high-quality trajectory and knowledge data to improve agent performance in target domains?} Traditional annotation methods suffer from subjectivity and prohibitive costs, limiting the scalability of GUI agent development.

To address this challenge, we introduce a self-evolving training pipeline centered on the Calibrated Step Reward System (CSRS). While prior work explores self-improvement through data generation via rationalization, self-critique, quality filtering, and offline reinforcement learning~\cite{zelikman2022star, madaan2023self_selfrefine, jung2023impossibledistill, gulcehre2023reinforced_rest}, these approaches primarily target single-turn tasks where model-generated annotations can be reliably verified. 
For multi-turn GUI agents where task execution spans multiple steps, direct model-generated annotations are prone to factual errors and hallucinations in the absence of verifiable ground truth. 
Unlike existing methods that rely solely on model introspection or post-hoc filtering, CSRS anchors LLM-generated dense step-level reasoning to trajectory-level evaluation signals, either automated verification scripts or human annotations, thereby ensuring annotation quality while maintaining scalability. 
Through trajectory-level calibration and LLM-powered knowledge extraction, CSRS converts model-generated trajectories into high-quality training data, achieving $>$90\% annotation accuracy with 10-100$\times$ cost reduction compared to traditional step-level annotation. Our progressive three-stage training paradigm orchestrates parallel data flows for novel exploration and strategic knowledge filtering, continuously improving model capabilities across multiple training rounds. 

On top of this pipeline, we train \textbf{Step-GUI}, a family of GUI specialist models (4B, 8B) built upon the Qwen3-VL backbone, and Step-GUI-8B achieves state-of-the-art performance: 80.2\% on AndroidWorld, 48.5\% on OSWorld, 65\% on ScreenShot-Pro, and 89.91\%/52.50\% on our AndroidDaily benchmark (static/end-to-end). Remarkably, our compact 4B model delivers competitive performance while remaining deployable on consumer-grade hardware, enabling local execution without cloud dependencies.

As GUI agents gain enhanced capabilities in visual understanding and autonomous task execution, two fundamental challenges emerge: \textit{standardized communication across heterogeneous devices} and \textit{user privacy protection when processing sensitive data}. The fragmentation of development efforts and high integration costs historically suppress innovation in tool ecosystems~\cite{qu2025toollearning}. To address the standardization challenge, the Model Context Protocol (MCP)~\cite{mcp} is proposed as a universal standard that defines how agents and tools should communicate, enabling an interoperable ``plug-and-play'' ecosystem. Recent work explores how agents can effectively retrieve and select appropriate MCP services from large pools~\cite{gan2025rag_ragmcp} and how to convert existing software projects into MCP format~\cite{ouyang2025code2mcp}. Building on these foundations, we propose \textbf{GUI-MCP} (Graphical User Interface - Model Context Protocol), the first MCP implementation specifically designed for GUI automation that addresses both standardization and privacy protection. GUI-MCP provides a standardized, cross-platform protocol that abstracts device capabilities into a small set of atomic and composite tools. Its hierarchical dual-layer architecture combines a \textit{Low-level MCP} that offers fine-grained operations (e.g., clicks, swipes, text input) with a \textit{High-level MCP} that delegates entire tasks to locally deployed GUI specialist models such as Step-GUI-4B. This design allows the main LLM to focus on high-level planning while it offloads routine GUI operations to local models. Critically, GUI-MCP supports a high-privacy execution mode in which raw screenshots and sensitive states stay on the device and only semantic summaries flow to external LLMs, which effectively protects user privacy while it leverages cloud-based reasoning capabilities.

As the community makes progress on these technical challenges, attention is shifting toward a more pressing concern: \textit{how to evaluate whether GUI agents can handle real-world everyday usage beyond synthetic setups?} Existing work develops diverse evaluation benchmarks, ranging from static benchmarks~\cite{cheng2024seeclick, wu2024atlas, li2025screenspotpro, li2025autogui, burns2022dataset_motif, rastogi2021uibert_refexp, zhang2024llamatouch, liu2024visualwebbench_vwb} that assess element grounding and task planning through prediction-annotation matching, to interactive benchmarks~\cite{rawles2024androidworld, xie2024osworld} that enable task execution in realistic operating system environments. However, these efforts either concentrate on single-step prediction accuracy or lack coverage of high-frequency Chinese mobile applications, leaving a critical gap: \textit{can agents reliably handle the high-frequency, everyday tasks that constitute real-world mobile usage?} To address this gap, we introduce \textbf{AndroidDaily}, a benchmark explicitly grounded in empirical analysis of authentic mobile usage patterns. Rather than pursuing maximal application coverage, AndroidDaily focuses on ubiquitous daily scenarios (Transportation, Shopping, Social Media, Entertainment, Local Services) where agent deployment has immediate practical impact. The benchmark employs a two-tier evaluation strategy: a \textit{Static Benchmark} with 3146 actions for efficient single-step action prediction, and an \textit{End-to-End Benchmark} with 235 tasks that span multiple dimensions (scenario, task type, complexity, and ambiguity) and evaluate autonomous task completion in fully functional environments. Step-GUI-8B demonstrates strong performance on AndroidDaily, which highlights both its current capabilities and remaining challenges in realistic deployment scenarios.

In summary, our contributions establish a comprehensive framework for building practical GUI agents:
\begin{itemize}
\item A self-evolving training pipeline with CSRS that achieves 10-100$\times$ cost reduction while it continuously improves model capabilities through closed-loop data refinement.
\item Step-GUI models that achieve state-of-the-art performance across multiple benchmarks (8B: 80.2\% on AndroidWorld, 48.5\% on OSWorld, 62.6\% on ScreenShot-Pro, 89.91\%/ 52.50\% on AndroidDaily static/end-to-end), with the 4B variant that enables consumer-grade local deployment.
\item GUI-MCP, the first standardized protocol for LLM-device interaction that balances execution efficiency with privacy protection through hierarchical architecture.
\item AndroidDaily, an ecologically valid benchmark grounded in real-world usage patterns with dual evaluation modes (3146 static actions, 235 end-to-end tasks) and multi-dimensional taxonomies for targeted capability analysis.
\end{itemize}

Together, these contributions address the complete pipeline from model training to standardized deployment interfaces and authentic evaluation, which paves the way for GUI agents that can assist users in their daily mobile interactions.

\section{Step-GUI}

\subsection{Data}
\subsubsection{Mid-train Data}

We aim to build a multimodal foundation model that retains \textbf{general-purpose capabilities}, inherits rich world knowledge, and acquires \textbf{strong agentic competence}. Instead of training a specialized planner or a domain-restricted policy model, we develop a multimodal foundation model capable of understanding diverse visual environments, following complex protocols, and executing multi-step tasks.

To bridge generic pretrained models and agent-specific training, we introduce a mid-train stage that equips the model with \textbf{foundational agent capabilities}. Specifically, the mid-trained model should: (i) retain and consolidate broad knowledge, including multimodal understanding, commonsense reasoning, and world knowledge; (ii) acquire fundamental visual competencies such as visual grounding, dense captioning, and fine-grained region understanding essential for GUI-based interactions; (iii) parse agent-style formats, including trajectory representations, action schemas, and observation structures; (iv) form initial instruction-to-action mappings that ground natural-language instructions into executable action sequences.

We construct a mid-train data mixture that balances general knowledge retention with agent-specific skill acquisition, comprising two complementary categories:

\begin{itemize}
    \item \textbf{General Multimodal \& Knowledge Data}, which preserves broad capabilities and builds foundational visual understanding:
    \begin{itemize}
        \item High-quality text and multimodal data (1.9M): curated general-purpose samples to maintain language fluency and multimodal reasoning;
        \item Knowledge-intensive data (2M): including dense captions and structured knowledge to reinforce world understanding;
        \item Grounding data (2.7M): region-level annotations for visual grounding, enabling precise localization required by GUI agents.
    \end{itemize}
    
    \item \textbf{Agent-Oriented Data}, which introduces action understanding and trajectory reasoning:
    \begin{itemize}
        \item Action alignment data (170K): samples that align natural-language instructions with atomic actions, bootstrapping the instruction-to-action mapping;
        \item Trajectory data (4M): multi-step interaction sequences annotated under diverse protocols by different annotators, exposing the model to varied action formats and planning patterns;
        \item Environment-specific data (420K): incorporating cross-platform trajectory samples from Android, Ubuntu, Windows, and macOS systems, offering comprehensive agent experiences across heterogeneous mobile and desktop environments.
    \end{itemize}
\end{itemize}
Through this balanced data mixture, the mid-trained model consolidates broad world knowledge, acquires essential visual competencies, learns to parse agent-style formats, and forms initial instruction-to-action mappings.

\subsubsection{Cold-start Data}

Building upon the mid-trained model's foundational agent capabilities, including visual grounding, agent-style format understanding, and initial instruction-to-action mappings, the cold-start stage focuses on knowledge injection and execution refinement. Rather than continuing broad exposure, this stage addresses knowledge gaps that cause execution failures and refines agent behavior through curated trajectory data. Specifically, the cold-start stage: (i) patches knowledge gaps by injecting missing knowledge identified from trajectory execution errors, enabling the model to overcome failure modes; (ii) refines agent behavior using curated trajectory samples to shape execution patterns; and (iii) preserves general competence by maintaining multimodal reasoning and visual grounding abilities. Agent failures often stem from knowledge deficiencies rather than insufficient behavioral examples. We therefore adopt an error-driven knowledge injection strategy: diagnosing execution failures and converting the missing knowledge into VQA pairs to directly target the model's weaknesses. The trajectory data serves as a behavioral scaffold, aligning outputs with agent formats while the enriched knowledge base enables robust generalization.

The cold-start data mixture (\textasciitilde1.67M samples) comprises:
\begin{itemize}
    \item \textbf{Knowledge Data} (864K, 52\%): 
          constructed by analyzing execution errors from trajectory rollouts. 
          When the model fails during trajectory execution, we identify the underlying missing knowledge (e.g., UI semantics, application behaviors, domain facts) and convert it into VQA-format samples. 
          This targeted injection directly addresses the model's knowledge blind spots rather than providing generic world knowledge;
    \item \textbf{Trajectory Data} (404K, 24\%): 
          high-quality multi-step interaction sequences serving as behavioral demonstrations;
    \item \textbf{General Multimodal Data} (284K, 17\%): 
          high-quality samples to retain broad multimodal reasoning capabilities;
    \item \textbf{Grounding Data} (122K, 7\%): 
          curated localization samples essential for GUI-based interactions.
\end{itemize}

Table~\ref{tab:data-comparison} presents the data composition across mid-train and cold-start stages. While mid-train employs a large-scale mixture (\textasciitilde11.2M samples) covering diverse data types including action alignment and environment-specific data, cold-start adopts a more focused mixture (\textasciitilde1.67M samples) with a higher proportion of knowledge data (52\%) to address identified execution failures:
\begin{table}[h]
\centering
\caption{Data composition comparison between mid-train and cold-start stages.}
\label{tab:data-comparison}
\begin{tabular}{lcc}
\toprule
\textbf{Data Type} & \textbf{Mid-Train} & \textbf{Cold-Start} \\
\midrule
General Multimodal & 1.9M & 284K \\
Grounding & 2.7M & 122K \\
Knowledge & 2.0M & 864K \\
Action Alignment & 170K & -- \\
Trajectory & 4.0M & 404K \\
Environment-Specific & 420K & -- \\
\midrule
\textbf{Total} & \textbf{\textasciitilde11.2M} & \textbf{\textasciitilde1.67M} \\
\bottomrule
\end{tabular}
\end{table}

The cold-start stage uses a smaller data volume (\textasciitilde1.67M vs. \textasciitilde11.2M), reflecting a shift from broad exposure to focused refinement. 
This stage emphasizes knowledge-driven optimization: identifying and patching execution failures through error-driven VQA knowledge injection, while using curated trajectory data to refine interaction patterns and execution behaviors.

\subsubsection{Grounding Data}

While grounding is traditionally framed as a perception-language alignment problem, this formulation becomes insufficient in the GUI domain, where the model is expected to understand and act within structured virtual environments. A graphical interface is effectively a microcosm of the world: it comprises entities, spatial hierarchies, causal affordances, and state transitions. Supervision that merely aligns text spans with visual regions fails to capture these richer semantics and often leads to brittle, appearance-driven behavior. This observation motivates a shift in how GUI grounding is formulated and trained. To succeed in GUI grounding, a model must move beyond surface-level visual matching and acquire capabilities analogous to a lightweight world model for virtual environments. In our formulation, this entails addressing three fundamental requirements:
\begin{itemize}
    \item \textbf{Functional semantics beyond appearance.} The model must learn that a gear icon affords settings and a trash icon affords deletion, rather than relying on surface-level visual similarity.
    \item \textbf{Latent world state.} The model should maintain a latent representation of what is visible, what is actionable, and how the interface state evolves under candidate actions.
    \item \textbf{World knowledge of HCI conventions.} Knowledge of human--computer conventions, layouts, and symbolic meanings enables reasoning about unseen or partially observed interfaces.
\end{itemize}

GUI grounding datasets face a fundamental challenge: annotations frequently contain errors (noise) and fail to accurately correspond to the semantic meaning of the interface elements they describe (misalignment). Directly scaling such data often amplifies noise rather than improving generalization. To address this, we design an iterative grounding-cleaning pipeline that progressively filters, corrects, and refines supervision using model feedback.

\begin{enumerate}
    \item \textbf{Initial grounding training.}
    We first train an initial model on raw open-source grounding data, general multimodal data, and knowledge-augmented annotations to establish basic perceptual alignment.

    \item \textbf{Pass-rate labeling with complexity scoring.}
    The trained model performs multiple independent rollouts per sample. Each sample receives a \textbf{pass-rate label} reflecting supervision quality. Additionally, an \textbf{LLM-based complexity scorer categorizes} tasks into simple localization, functional understanding, and intent-alignment levels, decoupling failures from annotation noise versus genuinely complex semantics.

    \item \textbf{Curriculum-based reliable data training.}
    High pass-rate samples serve as reliable supervision and are organized by complexity for difficulty-aware curriculum training: simple localization tasks stabilize early-stage grounding, while functional and intent-alignment tasks are progressively introduced via curriculum SFT and reinforcement learning.

    \item \textbf{Early exclusion of noisy cases.}
    Zero pass-rate samples which mix noisy annotations and genuinely hard cases, are excluded from early training to preserve learning signal quality.

    \item \textbf{Hard-case refinement.}
    Excluded samples are revisited at later stages: failed executions are rewritten with step-by-step knowledge and enriched annotations, then reintroduced as high-quality supervision.
\end{enumerate}

Through this closed-loop process of pass-rate labeling, data filtering, curriculum learning, and hard-case refinement, the model gradually transitions from sparse click-based supervision to explicit modeling of interface structure, dynamics, and affordances. This grounding loop substantially improves robustness on complex grounding and agentic tasks, while relying exclusively on open-source corpora and systematic scaling of model capacity, data volume, and compute.

\subsubsection{Trajectory Data}

\begin{figure}[t]
\centering
  \captionsetup{justification=justified, singlelinecheck=false}
\includegraphics[width=\linewidth]{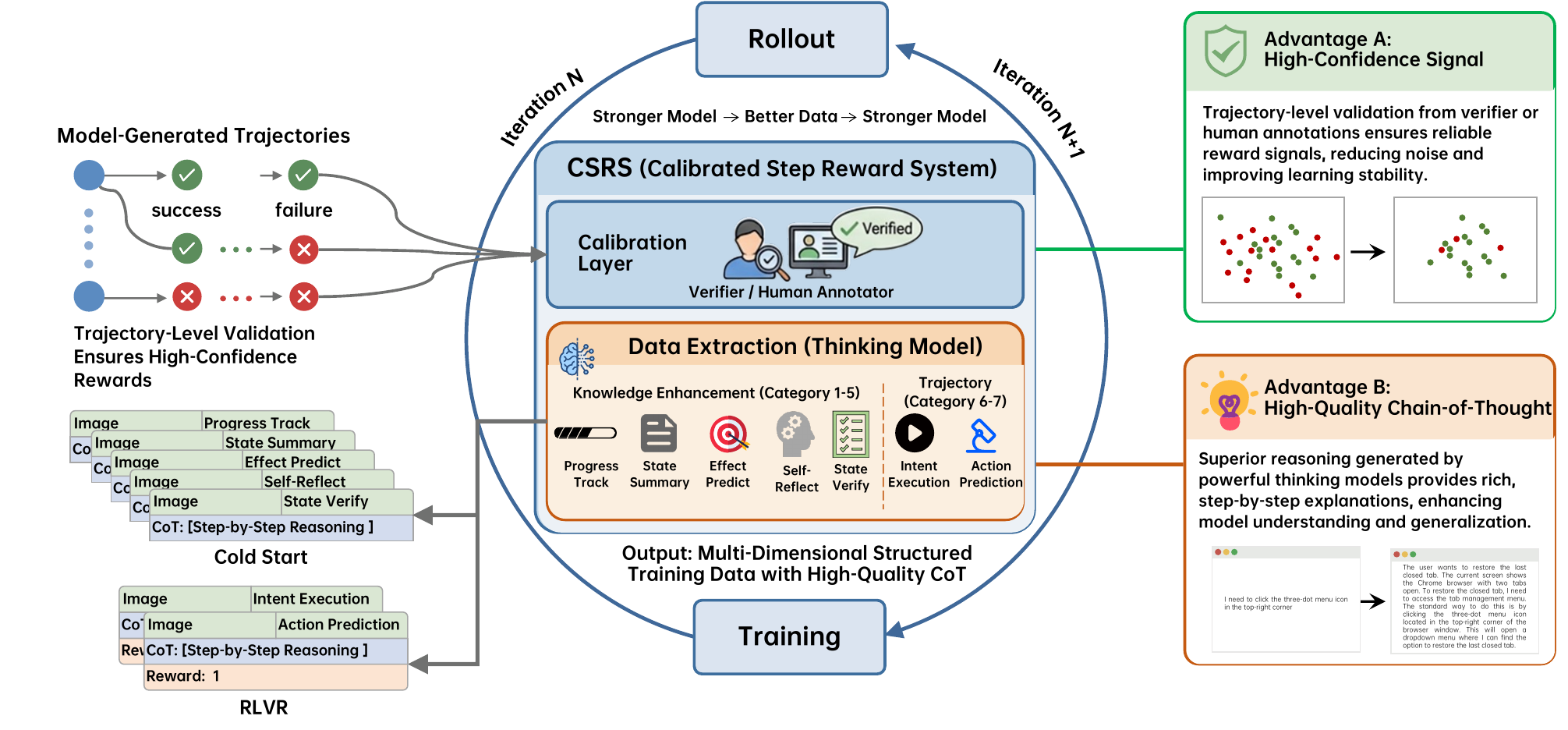}
\caption{\textbf{Calibrated Step Reward System (CSRS) Architecture.} The system consists of a Calibration Layer that performs trajectory-level validation (success/failure) and a Data Extraction module powered by thinking models that generates seven categories of structured training data. Model-generated trajectories flow through CSRS in an iterative loop: rollout generates trajectories, CSRS processes them into high-quality training data, and training produces stronger models for the next iteration. \textit{Advantage A}: Trajectory-level validation provides high-confidence reward signals, ensuring learning stability. \textit{Advantage B}: LLM-generated chain-of-thought provides rich reasoning that enhances model understanding. Success trajectories yield all seven data types while failed trajectories contribute only knowledge-related data (categories 1-6), implementing a selective learning strategy.}
\label{fig:csrs_architecture}
\end{figure}

To enable continuous model improvement through data flywheel iteration, we propose the Calibrated Step Reward System (CSRS), a novel data processing framework that converts model-generated trajectories into high-quality, multi-dimensional training data. CSRS serves as the critical bridge in the ``Rollout $\rightarrow$ CSRS $\rightarrow$ Training'' loop, ensuring data reliability while maximizing information extraction from each trajectory.
CSRS introduces two synergistic mechanisms that address fundamental challenges in reward system design (see Figure~\ref{fig:csrs_architecture}): (1) \textit{trajectory-level calibration} providing high-confidence reward signals, and (2) \textit{LLM-powered data extraction} generating superior chain-of-thought reasoning. Unlike traditional step-level annotation approaches that suffer from subjectivity and high costs, our trajectory-level validation achieves $>$90\% accuracy with 10-100$\times$ lower annotation costs by focusing on objectively verifiable task outcomes.

\textbf{System Architecture.}
As illustrated in Figure~\ref{fig:csrs_architecture}, CSRS consists of two main components. The \textit{Calibration Layer} employs verifiers or human annotators to perform binary success/failure validation at the trajectory level, establishing a reliable quality anchor. The \textit{Data Extraction Module}, powered by sophisticated thinking models, then generates seven categories of training data: (1) progress tracking, (2) state summary, (3) effect prediction, (4) self-reflection, (5) state verification, (6) intent execution, and (7) action prediction. This design ensures that all generated fine-grained data is anchored by high-confidence trajectory-level labels.

\textbf{Selective Learning Strategy.}
CSRS intelligently handles trajectories of different qualities. For successful trajectories, all seven data types are extracted, encompassing both knowledge enhancement (categories 1-5) and action prediction (categories 6-7). For failed trajectories, only knowledge-related data (categories 1-6) are retained, following the principle of ``learning knowledge from failures, but not learning erroneous actions.'' This selective strategy maximizes data utility while preventing the propagation of incorrect behaviors.

\textbf{LLM-Generated Knowledge Superiority.}
A key advantage of CSRS lies in leveraging powerful LLMs to automatically generate training data. Compared to human annotators, LLMs produce: (i) significantly richer chain-of-thought reasoning with detailed multi-step analysis, (ii) consistent quality across all samples without individual variation, (iii) comprehensive domain knowledge about GUI operations and application functionalities, and (iv) 80-90\% cost reduction through automation. For instance, while human annotators might simply label ``click center button'', CSRS generates detailed reasoning: ``The text is already selected. The next step is to apply center alignment formatting. I can see the alignment buttons in the toolbar, and I will click the 'Align Center' button. After clicking it, the heading should move to the center of the document.''

CSRS enables continuous model improvement through iterative training. In iteration $N$, model $M_n$ generates rollout trajectories, which are processed by CSRS to produce high-quality training data, resulting in an improved model $M_{n+1}$. As the model becomes stronger, rollout quality increases, leading to more successful trajectories and richer training data. This self-reinforcing cycle drives progressive performance enhancement: from initial success rates of 30-40\% to expert-level performance exceeding 85\% after multiple iterations.
The combination of trajectory-level calibration and LLM-powered generation achieves an optimal balance of quality, cost, and scalability. (\textit{Advantage A}) Trajectory-level validation from verifiers or human annotations ensures reliable reward signals, reducing noise in the learning process and improving training stability. (\textit{Advantage B}) Superior reasoning generated by powerful thinking models provides rich, step-by-step explanations, enhancing model understanding and generalization. This ``coarse-grained high-confidence labels + fine-grained high-quality content'' paradigm represents a significant advancement over traditional step-level annotation methods, establishing CSRS as a key infrastructure for building high-performance GUI agents.

\subsection{Training}
We employ the Qwen3-VL~\cite{Bai2025Qwen3VL} series (spanning 4B, 8B parameters) as our visual-linguistic backbone. 
To bridge the gap between general multimodal capabilities and expert GUI agency, we propose a progressive three-stage training paradigm: Mid-Training, Cold-Start Fine-Tuning, and Reinforcement Learning with Verifiable Rewards (RLVR).
\subsubsection{Formulation}
We formulate GUI automation as a sequential decision-making problem over the tuple $(\mathcal{Q}, \mathcal{A}, \mathcal{S}, \mathcal{O})$:

\begin{itemize}[leftmargin=*, itemsep=1pt]
    \item \textbf{Task Space $\mathcal{Q}$}: The set of natural language instructions that describe user intents (e.g., ``Check the battery level and send it to social media.''). Each task instance is denoted as $q \in \mathcal{Q}$.
    
    \item \textbf{Action Space $\mathcal{A}$}: GUI-based actions including click, scroll, swipe, type, and other interface interactions. These actions are executed purely through visual understanding without relying on structured accessibility information.
    
    \item \textbf{State Space $\mathcal{S}$}: The visual state of the GUI environment, represented by screen captures at each time step.
    
    \item \textbf{Observation Space $\mathcal{O}$}: Screenshots of the current interface along with the task instruction and historical interaction logs.
\end{itemize}

We model the agent as a parameterized policy $\pi_{\theta}: \mathcal{O} \times \mathcal{Q} \rightarrow \mathcal{A}$, where $\theta$ denotes the model parameters. Given a task $q \in \mathcal{Q}$ and visual observation $s \in \mathcal{S}$, the policy generates a sequence of outputs $o = (o_1, o_2, \ldots, o_T)$ representing the action trajectory. The objective is to optimize $\theta$ such that the policy successfully executes tasks through vision-based GUI interaction.

\begin{figure}[ht] 
  \centering
  \includegraphics[width=1\columnwidth]{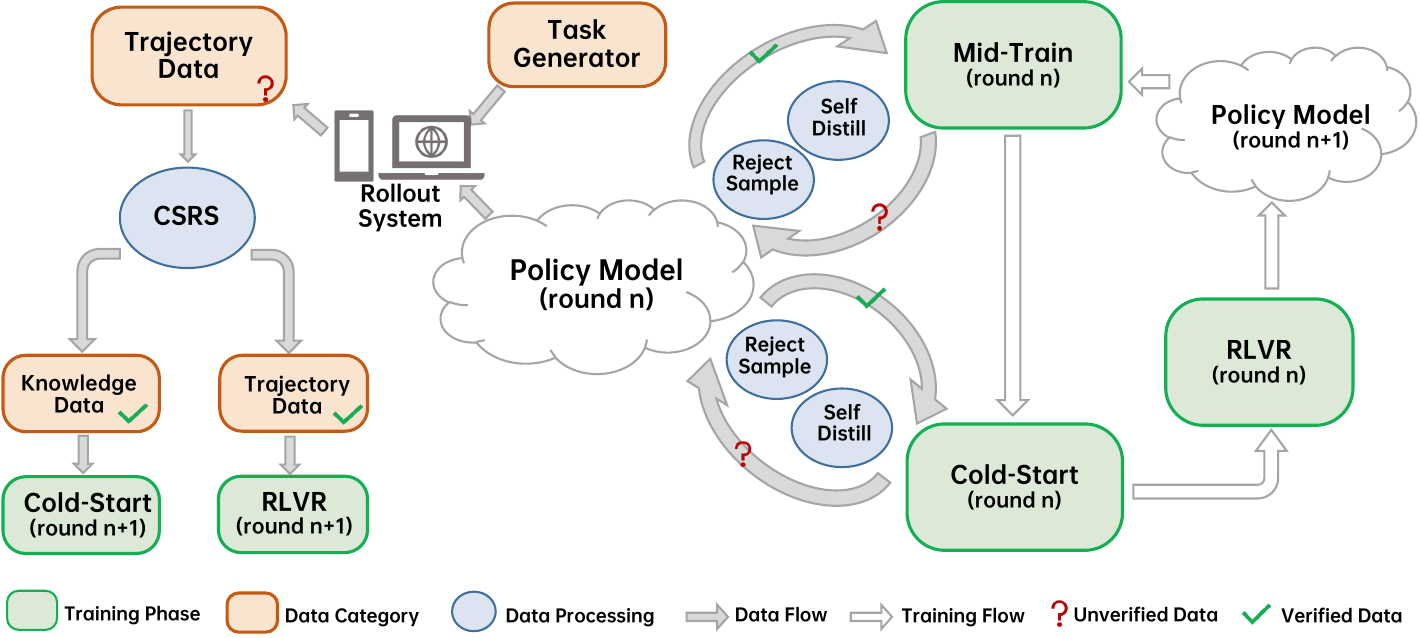} 
  \captionsetup{justification=justified, singlelinecheck=false}
  \caption{Self-Evolving Training Pipeline with Closed-Loop Data Refinement. The pipeline consists of three progressive training stages (Mid-Train, Cold-Start, and RLVR) and two parallel data flows. \textbf{Generation Data Flow:} The Policy Model generates new trajectories via Task Generator, which are verified through the CSRS to produce high-quality Knowledge Data and Trajectory Data for the next training round. \textbf{Refinement Data Flow:} Existing trajectory data undergo dual-path filtering through Self-Distillation and Rejection Sampling. This iterative loop continuously enhances data quality and model capability across rounds. }
 
  \label{fig:training} 
\end{figure}

\subsubsection{Data Flow}
Central to our approach is a self-evolving closed-loop data engine illustrated in Figure~\ref{fig:training}, which orchestrates two parallel data flows to continuously enhance both model capability and data quality across training rounds:

\noindent\textbf{Generation Data Flow.} The current Policy Model (round $n$) interactively executes newly generated tasks from the Task Generator within the Rollout System. During execution, the model produces raw trajectory data capturing its step-by-step interactions with the GUI environment. These raw trajectories are then processed by CSRS, which verifies action correctness and assigns calibrated rewards to filter and refine the data into two high-quality categories: (1) \textit{Knowledge Data} containing distilled task-solving insights and reasoning patterns, and (2) \textit{Trajectory Data} capturing verified complete multi-step execution paths. These high-quality synthetic data are channeled into Cold-Start and RLVR stages of the next round ($n+1$), enabling the model to learn from its own explorations.

\noindent\textbf{Refinement Data Flow.} In parallel, existing trajectory data undergo a dual-path filtering mechanism combining Self-Distillation and Rejection Sampling. This process partitions the data into two categories: (1) \textit{Accepted Set}---stable, high-confidence samples that consistently pass quality thresholds, recycled to Mid-Train and Cold-Start to reinforce foundational capabilities; and (2) \textit{Rejected Set}---challenging samples near the decision boundary that expose model weaknesses, exclusively routed to Cold-Start for targeted capability enhancement.

This dual-flow architecture ensures that each training round benefits from both \textit{novel high-quality exploration} (Generation Data Flow) and \textit{strategic refinement of existing knowledge} (Refinement Data Flow). The iterative cycle progressively amplifies the model's grounding accuracy, general multimodal understanding, and complex reasoning capabilities in GUI scenarios.

\subsubsection{Reinforcement Learning with Verifiable Rewards}
To specialize the agent in precise reasoning and execution within the GUI domain, we employ Group Relative Policy Optimization (GRPO). The objective function is defined as:
\begin{equation}
J_{GRPO}(\theta) = \mathbb{E}_{\substack{q \sim P(Q), \{o_i\}_{i=1}^G \sim \pi_{\theta_{old}}(O|q)}} \left[ \frac{1}{G} \sum_{i=1}^G \frac{1}{|o_i|} \sum_{k=1}^{|o_i|} \min \left( r_{i,k} A_i, \text{clip}(r_{i,k}, 1-\epsilon, 1+\epsilon) A_i \right) - \beta \mathbb{D}_{KL} \right]
\label{equ: grpo}
\end{equation}
\noindent where $G$ is the group size, $A_i$ is the advantage computed by normalizing rewards within the group, $q$ is the task sampled from $P(Q)$, $\pi_{\theta_{old}}$ is the reference policy, $o_i$ is the trajectory generated by $\pi_{\theta_{old}}$, $r_{i,k} = \frac{\pi_{\theta}(o_{i,k}|q,o_{i,<k})}{\pi_{\theta_{old}}(o_{i,k}|q,o_{i,<k})}$ is the importance sampling ratio, $\epsilon$ is the clipping parameter, and $\beta$ controls the KL divergence regularization.

\noindent\textbf{Fine-Grained Hybrid Reward Specification.}
We construct a composite reward function $R(o, s)$ by integrating three distinct signal categories: verifiable spatial metrics, action-semantic validity, and model-based capability assessment.

\paragraph{1) Spatial-Geometric Dense Rewards.}
To ensure pixel-level precision, we implement a dense reward mechanism based on tolerance-normalized high-order decay. Let $\tau_x, \tau_y$ denote the predefined pixel-level tolerance thresholds.

\noindent\textit{Point-based Reward:} For coordinate interactions (e.g., grounding, CLICK, LONGPRESS), we define the normalized deviation $\hat{\delta}_k = |k_{pred} - k_{gt}| / \tau_k$. The reward $r_{point}$ employs a quartic decay function to impose steep penalties for deviations beyond the tolerance zone:
\begin{equation}
    r_{point} = \exp \left( - \left( \hat{\delta}_x^4 + \hat{\delta}_y^4 \right) \right)
\end{equation}

\noindent\textit{BBox-based Reward:} For region grounding, we decouple the bounding box into center coordinates $(c_x, c_y)$ and dimensions $(w, h)$. We construct a composite geometric energy term $E_{geom}$ that down-weights dimensional errors by a factor of $\lambda$:
\begin{equation}
    E_{geom} = \hat{\delta}_{cx}^4 + \hat{\delta}_{cy}^4 + \lambda \hat{\delta}_w^4 + \lambda \hat{\delta}_h^4
\end{equation}
where $\lambda=0.5$. To simultaneously optimize for alignment accuracy and area overlap, the final bounding box reward $R_{bbox}$ is a weighted fusion of the geometric score and Intersection over Union (IoU):
\begin{equation}
    R_{bbox} = \alpha \cdot \exp(-E_{geom}) + (1 - \alpha) \cdot \text{IoU}(b_{pred}, b_{gt})
\end{equation}
where we set $\alpha = 0.8$ to prioritize geometric centrality while retaining IoU as a shape-consistency regularization.

\paragraph{2) Action-Semantic Mixed Rewards.}
For the GUI action space, we decouple the reward signal into action type and action value.
\begin{itemize}[leftmargin=*, itemsep=1pt]
    \item \textbf{Sparse Action Type:} The validity of the chosen operation (e.g., WAIT vs. COMPLETE) is modeled as a binary sparse reward $\mathbb{I}(\hat{a}_{type} = a^*_{type})$.
    \item \textbf{Adaptive Value Modeling:} The reward for action parameters varies by type. For trajectory-based vectors (e.g., SLIDE), we compute the cosine similarity between the predicted vector $\mathbf{v}_{pred}$ and the ground truth $\mathbf{v}_{gt}$, mapping the alignment to a dense $[0, 1]$ interval. For semantic actions requiring information retrieval (e.g., INFO, TYPE), we utilize an external LLM to verify the content, returning a scalar score $s \in [0, 1]$.
\end{itemize}

\paragraph{3) Soft Capability Rewards (LLM-as-a-Judge).}
For abstract qualities where deterministic rules are inapplicable, we employ an LLM-as-a-Judge mechanism. 
This module evaluates the generated trajectory based on intent consistency, fluency, and reasoning quality, providing a complementary soft signal that aligns the policy with human-preferred interaction patterns.

\noindent\textbf{Semi-Online Exploration with Hindsight.}
Exploration in long-horizon GUI tasks is notoriously difficult due to sparse rewards. To mitigate this, we introduce a Semi-Online training strategy. For rollout groups that fail to complete the task, we inject Ground-Truth Hints into the prompt to guide the model through the correct reasoning path during a second pass. This allows the model to experience high-reward trajectories that were previously out of its capability range, effectively converting "failed explorations" into "guided successful samples" with high advantage scores.

\noindent\textbf{Stability \& Efficiency Enhancements.}
To ensure stable convergence and maximize data utility, we integrate several algorithmic enhancements:
\textbf{1) Dynamic Exploration ($\epsilon_{high}$):} We introduce a dynamic parameter $\epsilon_{high}$ to modulate the clipping range, providing greater flexibility for low-probability actions. This expands the exploration space without destabilizing the policy update.
\textbf{2) Sample Reuse via Importance Sampling:} As illustrated in the efficiency module, data generation is computationally expensive. We employ Importance Sampling to reuse collected trajectories for multiple gradient updates. The policy is updated for $K$ iterations per rollout batch. 
The importance sampling ratio $r_t(\theta) = \frac{\pi_\theta(o_t|q,o_{<t})}{\pi_{old}(o_t|q,o_{<t})}$ accounts for the distributional shift between the current evolving policy and the data-collecting policy, significantly improving sample efficiency while maintaining trust-region constraints.

\noindent\textbf{Gradient Preservation for Clipping.} The gradient of Equation~\eqref{equ: grpo} without $\mathbb{D}_{KL}$ can be formulated as:
\begin{equation}
\begin{aligned}
\nabla_\theta J_{GRPO}(\theta) & = \mathbb{E}_{\substack{q \sim P(Q),  \{o_i\}_{i=1}^G \sim \pi_{\theta_{old}}(O|q)}} \left[ \frac{1}{G} \sum_{i=1}^G \frac{1}{|o_i|} \sum_{k=1}^{|o_i|}f(r_{i,k}, {A}_{i,k})\nabla_\theta \log \pi_{\theta}(o_{i,k} | q, o_{i, <k}) \right],\\
&\text{where } f(r_{i,k}, {A}_{i,k}) = \begin{cases}
A_i \cdot r_{i,k} & \text{if } 1-\epsilon < r_{i,k} < 1+\epsilon, \\
\beta_1 \cdot A_i \cdot (1+\epsilon) & \text{if } r_{i,k} > 1+\epsilon, \\
\beta_2 \cdot A_i \cdot (1-\epsilon) & \text{if } r_{i,k} < 1-\epsilon,
\end{cases}\\
\label{equ: gp_gradient}
\end{aligned}
\end{equation}
where the gradient modulation depends explicitly on the importance sampling ratio $r_{i,k}$. 
A critical analysis of Equation~\eqref{equ: gp_gradient} reveals an inherent instability for tokens with low baseline probabilities. For tokens where the reference probability $\pi_{\theta_{\text{old}}}$ is minimal, even marginal positive perturbations in the current policy $\pi_{\theta}$ result in a disproportionate amplification of the ratio $r_{i,k}$. Consequently, these low-probability tokens are highly susceptible to ratio explosion, prematurely triggering the upper clipping threshold $(1+\epsilon)$. 
In standard clipping mechanisms, this saturation effectively zeros out the gradient, preventing the model from reinforcing potentially valuable but currently rare actions, a phenomenon that severely hinders the exploitation of sparse rewards. 
To mitigate this issue and maintain effective optimization, we implement a Gradient Preservation strategy~\cite{su2025gppo}.
As defined in the piecewise function $f(r_{i,k}, {A}_{i,k})$, we retain a scaled gradient even when the ratio exceeds the trust region, thereby ensuring continuous learning signal flow for low-probability tokens.

\section{GUI-MCP: An Efficient Protocol for Secure LLM-Device Interaction}
\label{sec:gui-mcp}

\begin{figure}[ht] 
  \centering
  \captionsetup{justification=justified, singlelinecheck=false}
  \includegraphics[width=0.98\columnwidth]{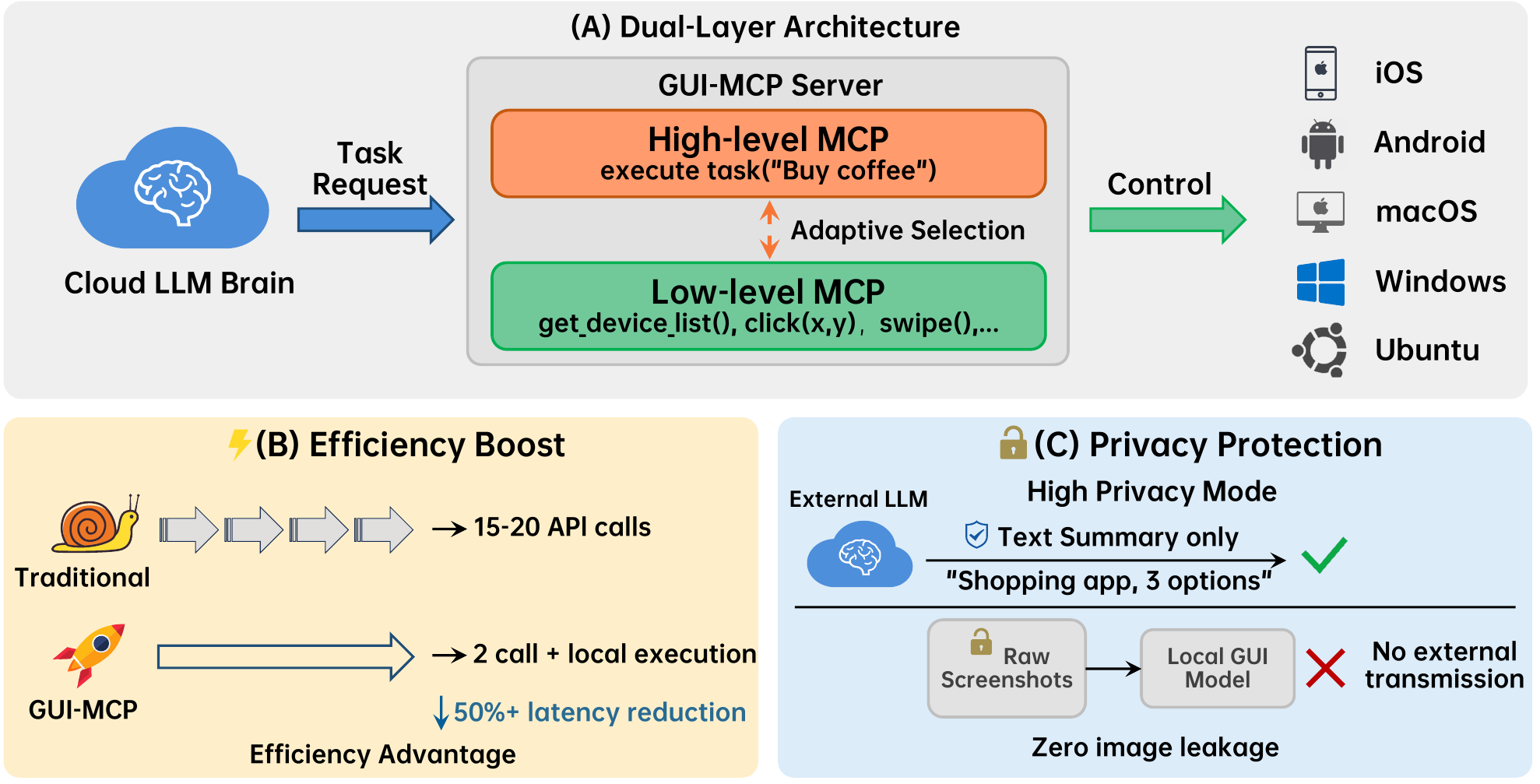} 
  \caption{Overview of GUI-MCP architecture. The dual-layer design includes Low-level MCP (providing atomic device operations) and High-level MCP (delegating tasks to a local GUI specialist model). This hierarchical approach enables efficient task execution while preserving user privacy through local processing.}
  \label{fig:gui-mcp-architecture} 
\end{figure}

Despite the remarkable progress in LLMs, their application to GUI automation remains challenging due to the lack of standardized interfaces for cross-platform device control. Existing solutions are often platform-specific and require substantial engineering effort to integrate with different LLMs and devices. To address this limitation, we propose \textbf{GUI-MCP} (Graphical User Interface - Model Context Protocol), the first MCP implementation specifically designed for GUI operation tasks. As shown in Figure~\ref{fig:gui-mcp-architecture}, GUI-MCP provides a standardized toolkit that seamlessly connects various LLMs with multiple device platforms (Ubuntu, macOS, Windows, Android, iOS), enabling LLMs to control mobile and desktop devices through a unified protocol for executing GUI operation tasks.

\subsection{Architecture Design}

GUI-MCP adopts a hierarchical design that stratifies functionality into two distinct levels: \textbf{Low-level MCP} and \textbf{High-level MCP}, as illustrated in Figure~\ref{fig:gui-mcp-architecture}.

\subsubsection{Low-level MCP}
The low-level MCP focuses on atomic device operations, providing fine-grained control interfaces. This layer exposes the following categories of primitives:

\noindent\textbf{Device Management.} The interface \texttt{get\_device\_list()} retrieves all connected devices, enabling multi-device orchestration.

\noindent\textbf{State Perception.} The interface \texttt{get\_screenshot()} captures the current device screen state, providing visual feedback for decision-making.

\noindent\textbf{Basic Operations.} A comprehensive set of interaction primitives as shown in Table~\ref{tab:low-level-mcp}.

\begin{table}[h]
\centering
\small
\begin{tabular}{ll}
\hline
\textbf{Category} & \textbf{Operations} \\
\hline
Click & \texttt{click}, \texttt{double\_click}, \texttt{triple\_click} \\
      & \texttt{right\_click}, \texttt{middle\_click} \\
\hline
Gesture & \texttt{swipe}, \texttt{long\_press} \\
\hline
Mouse & \texttt{move\_to}, \texttt{drag\_to} \\
\hline
Input & \texttt{input\_text}, \texttt{hotkey} \\
\hline
Application & \texttt{awake} \\
\hline
\end{tabular}
\caption{Low-level MCP basic operations.}
\label{tab:low-level-mcp}
\end{table}

These atomic interfaces provide maximum flexibility for the main LLM to perform fine-grained planning and control based on the current state and task requirements.

\subsubsection{High-level MCP}
The high-level MCP focuses on abstract task execution by encapsulating complete task execution logic. The primary interface is:

\noindent\texttt{execute\_task(task\_description)}: This interface accepts natural language task descriptions and automatically completes the task. For example:
\begin{itemize}
    \item \texttt{execute\_task("Click the first element")}
    \item \texttt{execute\_task("Buy a cup of coffee")}
    \item {\small\texttt{execute\_task("Search for white canvas shoes,\\size 37, under \$100, and favorite the first result")}}
\end{itemize}

Internally, the high-level MCP integrates a locally deployed GUI specialist model (e.g., Step-GUI-4B) that has been specifically optimized for GUI operation tasks. This specialist model can autonomously complete tasks within its capability scope. The system prompt of the main LLM explicitly describes the capability boundaries of the GUI specialist model, helping the main LLM determine when to delegate tasks to the high-level MCP.

\subsection{Design Advantages}

\subsubsection{Improved Execution Efficiency}
The dual-layer architecture enables the main LLM to flexibly select control strategies based on task complexity and current state:

\noindent\textbf{Low-level MCP Usage Scenarios:}
\begin{itemize}
    \item Rapid acquisition of current device state
    \item Tasks requiring fine-grained step-by-step planning
    \item Tasks exceeding the GUI specialist model's capabilities
    \item Scenarios requiring multi-turn user interaction for task clarification
\end{itemize}

\noindent\textbf{High-level MCP Usage Scenarios:}
\begin{itemize}
    \item Clear task descriptions within the GUI specialist model's scope
    \item Desire to reduce inference overhead and API calls to the main LLM
    \item Tasks with strong independence that can be completed in one shot
\end{itemize}

By appropriately allocating tasks, the main LLM can delegate simple, repetitive GUI operations to the local specialist model while focusing on high-level task planning and decision-making, thereby significantly improving overall execution efficiency.

\subsubsection{Enhanced Privacy Protection}
In an era where privacy protection is increasingly critical, many users have concerns about transmitting screenshots and device information to external cloud LLM service providers. GUI-MCP provides a \textbf{High Privacy Mode} with the following characteristics:

\noindent\textbf{Data Anonymization Mechanism:} External cloud LLM brains cannot directly access raw screenshots and detailed device information. Instead, they only receive state summaries processed by the local GUI model. These summaries contain only the key semantic information necessary for task completion, excluding sensitive visual details.

\noindent\textbf{Local Execution:} Operations that require screenshot analysis for planning are performed by the local GUI model. All image data and sensitive information are processed exclusively on the local device, with the external cloud LLM only responsible for high-level task decomposition and decision-making.

\noindent\textbf{Flexible Privacy Levels:} Users can configure privacy protection levels based on their trust preferences and task requirements. The system supports multi-level configurations ranging from fully open (direct screenshot transmission) to fully private (text summaries only).

This design allows GUI-MCP to fully leverage the reasoning capabilities of powerful cloud-based LLMs while effectively protecting user privacy data, achieving an optimal balance between functionality and privacy.

Through its innovative dual-layer architecture, GUI-MCP provides a systematic solution for LLM-driven GUI automation across both efficiency and privacy dimensions. This protocol not only lowers the technical barrier for extending LLM capabilities to the GUI operation domain but also provides a viable path for building AI assistants that are both powerful and privacy-preserving.

\section{AndroidDaily: A Dynamic Benchmark for Agentic Tasks in Daily Life}
\label{bmk_AndroidDaily}

\begin{figure}
    \centering
    \captionsetup{justification=justified, singlelinecheck=false}
    \includegraphics[width=1.0\linewidth]{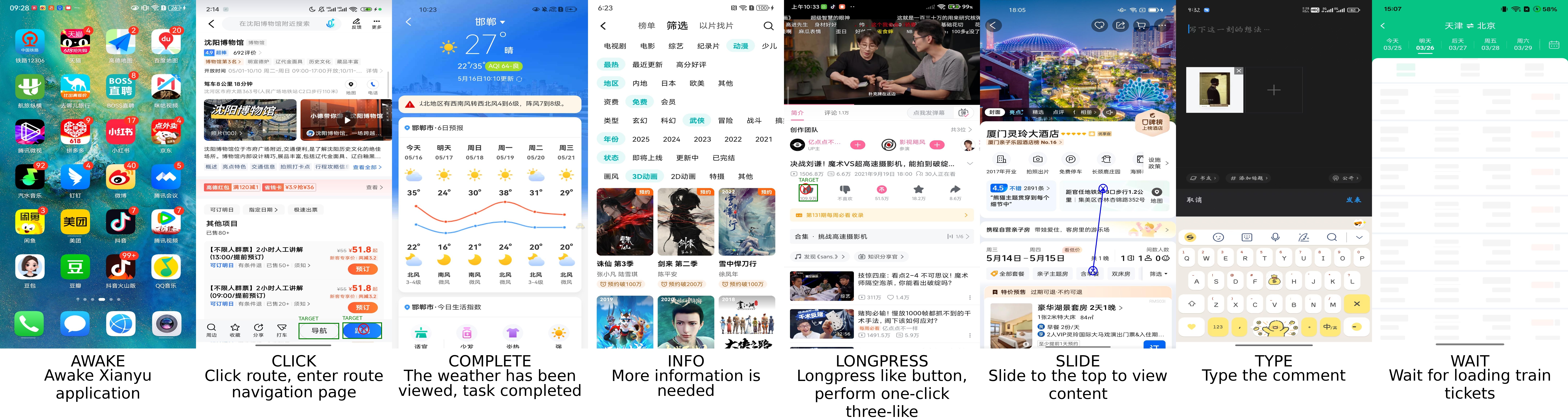}
    \caption{AndroidDaily Static Benchmark Action Taxonomy. Eight action types for Android task automation are illustrated: AWAKE, CLICK, COMPLETE, INFO, LONGPRESS, SLIDE, TYPE, and WAIT (left-to-right, top-to-bottom). Each example shows a task description and annotated ground truth actions with parameters. Multi-solution cases are supported, e.g., the CLICK example (second panel) shows two valid target regions highlighted in red boxes.}
    \label{fig:android_daily_static}
\end{figure}

While recent benchmarks like AndroidWorld make significant strides in evaluating GUI agents on Android platforms, a critical gap remains: the disconnect between evaluated tasks and actual daily usage patterns. Existing benchmarks predominantly focus on productivity or utility apps that, while technically diverse, do not necessarily represent the tasks users perform most frequently in their everyday lives. AndroidDaily addresses this gap by grounding task selection in empirical analysis of actual mobile usage patterns rather than application catalogs. Based on usage frequency data and download statistics, we curate apps spanning key daily-life scenarios including transportation, shopping, social media, entertainment, and local services, prioritizing high-frequency tasks such as food delivery ordering, ride-hailing, short-form video consumption, and mobile payments. The selected tasks involve real-world consequences (e.g., financial transactions, service bookings), multi-step decision-making (e.g., comparing options with multiple criteria), and tight integration with physical services. These characteristics fundamentally differentiate AndroidDaily from utility-focused benchmarks, enhancing the representativeness of scenarios where agent deployment has immediate practical impact. AndroidDaily employs a dual evaluation strategy that balances evaluation efficiency with real-world fidelity, including a static benchmark and an end-to-end benchmark.

\begin{figure}
    \centering
    \captionsetup{justification=justified, singlelinecheck=false}
    \includegraphics[width=1\linewidth]{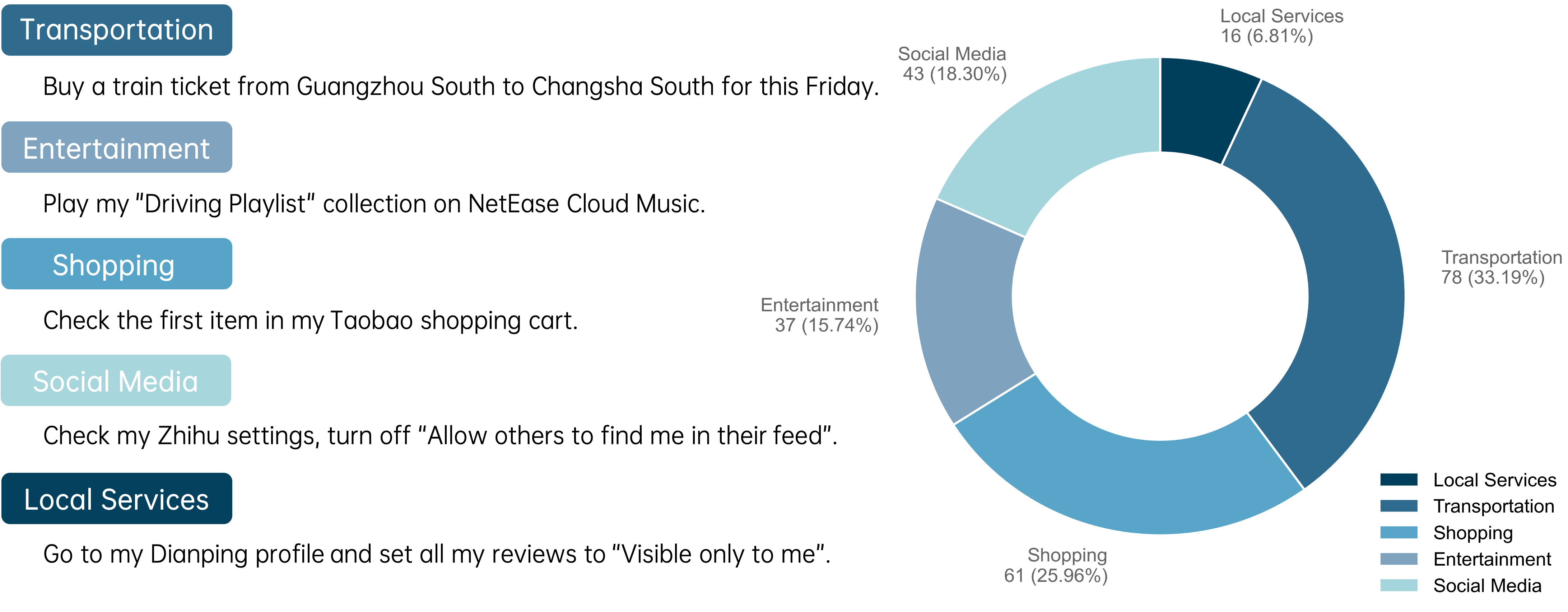}
    \caption{AndroidDaily End-to-End Benchmark Scenario Distribution. The 235 tasks are organized across five daily-life scenarios: Transportation, Shopping, Social Media, Entertainment, and Local Services. The pie chart (right) shows the quantitative distribution across these categories, with Transportation and Shopping comprising the majority of tasks. Representative task examples for each scenario are provided (left), demonstrating the diversity and practical relevance of the benchmark.}
    \label{fig:scenario_distribution}
\end{figure}

\paragraph{Static Benchmark (3146 actions).} The static benchmark provides task descriptions alongside step-by-step screenshots, evaluating agent performance by comparing predicted actions against ground-truth single-step actions. As illustrated in Figure~\ref{fig:android_daily_static}, the benchmark encompasses eight action types essential for Android task automation: \textit{AWAKE}, \textit{CLICK}, \textit{COMPLETE}, \textit{INFO}, \textit{LONGPRESS}, \textit{SLIDE}, \textit{TYPE}, and \textit{WAIT}. Each action is annotated with precise parameters that define execution targets and values. Importantly, the benchmark explicitly supports multi-solution ground truth annotations, for instance, CLICK actions may have multiple valid target regions when several UI elements can accomplish the same goal, reflecting the reality that mobile interfaces often offer alternative interaction pathways. We compute accuracy for both action types and action values. This approach offers efficient testing without requiring complex engineering infrastructure, enabling rapid iteration during model development.

\paragraph{End-to-End Benchmark (235 tasks).} The end-to-end benchmark evaluates agents on complete task workflows, employing an LLM-based judger to determine task completion and measuring overall success rates. Though more demanding in terms of infrastructure, this setup provides ecologically valid evaluation that closely mirrors real-world task execution scenarios.

As shown in Figure~\ref{fig:scenario_distribution}, the 235 end-to-end tasks are primarily organized by scenario distribution, spanning five core categories: \textit{Transportation} (78 tasks, 33.19\%), \textit{Shopping} (61 tasks, 25.96\%), \textit{Social Media} (43 tasks, 18.30\%), \textit{Entertainment} (37 tasks, 15.74\%), and \textit{Local Services} (16 tasks, 6.81\%). Beyond scenario categorization, tasks are further characterized by cognitive operation types (information filtering, querying, and analysis), structural complexity (atomic, composite, and conditional/loop tasks), and instruction ambiguity levels (low, medium, and high ambiguity). This multi-dimensional organization enables targeted analysis of agent capabilities and systematic identification of failure modes, offering actionable insights for model improvement.

\section{Experiments}

\subsection{Experimental Setup}

\noindent\textbf{Grounding Benchmarks.}
We evaluate the grounding capabilities of our model on five challenging benchmarks: ScreenSpot-Pro~\cite{li2025screenspotpro}, ScreenSpot-v2~\cite{wu2024atlas}, OSWorld-G~\cite{wu2024atlas},MMBench-GUI-L2~\cite{wang2025mmbench} and VisualWebBench~\cite{liu2024visualwebbench_vwb}. These benchmarks assess the model's ability to accurately perceive and locate UI elements across common GUI pages with varying levels of difficulty, providing a comprehensive evaluation of fundamental visual grounding skills in graphical user interfaces.

\begin{table}[t]
\centering
\captionsetup{justification=justified, singlelinecheck=false}
\caption{Model Performance on Multiple GUI-grounding Benchmarks. 
Best results are in \textbf{bold}, second best are \underline{underlined}.}
\label{tab:gui_benchmark_performance}
\resizebox{\linewidth}{!}{
\begin{tabular}{l|ccccc}
\toprule
\textbf{Model} 
& ScreenSpot-Pro 
& ScreenSpot-v2 
& OSWorld-G 
& MMBench-GUI-L2 
& VisualWebBench \\
\midrule

OpenAI CUA & 23.4 & 87.9 & 36.4 & - & - \\
UI-TARS-1.5 & \underline{61.6} & 94.2 & - & - & - \\
SeedVL-1.5 & 60.9 & \textbf{95.2} & - & \underline{84.4} & 87.3 \\
InternVL3.5-30B-A3B-Instruct & - & 87.3 & 42.4 & - & - \\

UI-TARS-1.5-7B & 35.7 & 91.6 & 47.5 & 64.3 & - \\
UI-TARS-72B-DPO & 38.1 & 90.3 & - & 74.2 & 82.8 \\
Qwen2.5-VL-72B & 43.6 & 87.1 & - & 41.8 & - \\

GUI-Owl-7B & 54.9 & 92.8 & 55.9 & 80.5 & - \\
GUI-Owl-32B & 58.0 & 93.2 & 58.0 & 83.0 & - \\

Qwen3-VL-4B-Instruct & 59.5 & 94.0 & 58.2 & - & - \\
Qwen3-VL-8B-Instruct & 54.6 & 94.4 & 58.2 & - & - \\
Qwen3-30B-A3B-Instruct & 60.5 & 94.7 & 61.0 & 79.7 & 79.7 \\

\textbf{Step-GUI-4B} & 60.0 & 93.6 & \underline{66.9} & 84.0 & \textbf{90.7} \\
\textbf{Step-GUI-8B} & \textbf{62.6} & \underline{95.1} & \textbf{70.0} & \textbf{85.6} & \underline{89.7} \\
\bottomrule
\end{tabular}
}
\end{table}

\noindent\textbf{Computer Use.} OSWorld~\cite{xie2024osworld} provides a comprehensive evaluation suite consisting of 369 real-world tasks with detailed environment configurations and automated evaluation scripts. These tasks cover diverse desktop applications and require multi-step reasoning and interaction capabilities.

\noindent\textbf{Phone Use.} AndroidWorld~\cite{rawles2024androidworld} offers 116 tasks across 20 mobile applications within a live Android emulator environment. The benchmark includes dynamic task variations generated via randomized parameters, enabling robust evaluation of agent performance.

\noindent\textbf{AndroidDaily.} As described in Section~\ref{bmk_AndroidDaily}, AndroidDaily evaluates agent performance on commonly used Chinese applications with tasks categorized into three difficulty levels (L1-L3), providing insights into model capabilities on region-specific mobile interfaces.

\noindent\textbf{General Multimodal Benchmarks.}
We further evaluated the generalizability of our model on mainstream benchmarks across other multimodal domains which contains V*~\cite{wu2024v}, PhyX~\cite{shen2025phyx}, OmniOCR~\cite{Omniocr}, OCRBench~\cite{liu2024ocrbench}, LogicVista~\cite{xiao2024logicvista}, MMBench\_en~\cite{liu2024mmbench}, MMBench\_cn, MMStar~\cite{chen2024we},MathVista~\cite{lu2023mathvista},CharXiv~\cite{wang2024charxiv} and Simplevqa~\cite{cheng2025simplevqa}.

\noindent\textbf{Baselines.}
We compare our approach against state-of-the-art foundation models and specialized GUI agents: Claude-4-sonnet~\cite{claude4sonnet}, Claude-4.5-sonnet~\cite{claude45sonnet}, OpenAI CUA-o3 (Computer Use Agent)~\cite{openai_cua_o3}, Gemini-2.5-Computer Use~\cite{gemini2.5cu}, UI-TARS-1~\cite{qin2025ui_uitars1}, UI-TARS-2~\cite{wang2025ui_uitars2}, InternVL3.5~\cite{wang2025internvl3}, MobileRL-9B~\cite{xu2025mobilerl}, AutoGLm-9B~\cite{AutoGLM}, GLM4.6V~\cite{glm4.6v}, Qwen2.5VL~\cite{bai2025qwen2}, Qwen-OWL\&Mobile-Agent-v3 ~\cite{ye2025mobile_mobileagentv3}, Qwen3VL~\cite{Bai2025Qwen3VL}. These baselines represent diverse paradigms including general-purpose vision-language models, specialized computer-use models, and dedicated GUI automation agents.

\subsection{Main Results}

\noindent\textbf{Performance on Grounding Benchmarks.}
Table~\ref{tab:gui_benchmark_performance} presents a comprehensive comparison of our method against state-of-the-art baselines across five GUI grounding benchmarks. Our models, Step-GUI-4B and Step-GUI-8B, demonstrate consistently strong performance across diverse evaluation settings.

\textbf{ScreenSpot-Pro}. On ScreenSpot-Pro, which evaluates professional-level GUI grounding, Step-GUI-8B achieves the highest score of 62.6, outperforming UI-TARS-1.5 (61.6), SeedVL-1.5 (60.9), and Qwen3-30B-A3B (60.5). Step-GUI-4B also attains a competitive 60.0, surpassing models with significantly larger parameter counts such as Qwen2.5-VL-72B (43.6).

\textbf{ScreenSpot-v2}. On ScreenSpot-v2, Step-GUI-8B reaches 95.1, approaching the top-performing SeedVL-1.5 (95.2) and exceeding UI-TARS-1.5 (94.2).

\textbf{OSWorld-G}. For OSWorld-G, which measures grounding in realistic desktop environments, our models achieve substantial improvements: Step-GUI-8B attains 70.0 and Step-GUI-4B reaches 66.9, outperforming the next best Qwen3-30B-A3B (61.0) by a significant margin of +9.0 and +5.9 points respectively.

\textbf{MMBench-GUI-L2}. On MMBench-GUI-L2, Step-GUI-8B achieves the best result of 85.6, followed by Step-GUI-4B at 84.0, both surpassing SeedVL-1.5 (84.4) and GUI-Owl-32B (83.0). 

\textbf{VisualWebBench}. For VisualWebBench, Step-GUI-4B achieves the highest score of 90.7, exceeding SeedVL-1.5 (87.3) by +3.4 points.

These results validate the effectiveness of our multi-stage training pipeline: mid-training establishes robust visual grounding and agent-style format parsing, cold-start knowledge injection addresses task-critical gaps, and the iterative grounding-cleaning pipeline ensures high-quality supervision throughout training. Notably, our compact 4B and 8B models consistently match or outperform substantially larger models (30B--72B), demonstrating strong parameter efficiency.

\noindent\textbf{Performance on End-to-End Benchmarks.}
\begin{table*}[t]
\centering
\captionsetup{justification=justified, singlelinecheck=false}
\caption{Performance on End-to-End Environment Benchmarks. Best results are in \textbf{bold}, second best are \underline{underlined}. Note: We report both pass@1 and pass@3 metrics. Pass@3 (maximum 3 attempts per task) is used to mitigate environment instability issues including occasional CAPTCHA verification, virtual machine crashes, and other non-model-related failures, providing a more reliable evaluation of actual model capabilities.} 
\label{tab:online_env_benchmarks}
\small %
\begin{tabular}{l|cc} 
\toprule
\textbf{Model} & OSWorld-Verified & AndroidWorld \\
\midrule
Claude-4-sonnet & 43.9 & - \\
Claude-4.5-sonnet & \textbf{61.4} & - \\
OpenAI CUA o3 & 23 & - \\
Gemini-2.5-Computer Use & - & 69.7 \\
SeedVL-1.5 & 34.1 & 62.1 \\
UI-TARS-1.5 & 42.5 & 64.2 \\
UI-TARS-2 & 47.5 & 73.3 \\
UI-TARS-1.5-7B & 27.4 & - \\
UI-TARS-72B-DPO & 24.0 & 46.6 \\
GUI-Owl-7B & 34.9 & 66.4 \\
Mobile-Agent-v3 & 37.7 & 73.3 \\
Qwen3-VL-4B-Instruct & 26.2 & 45.3 \\
Qwen3-VL-8B-Instruct & 33.9 & 47.6 \\
Qwen3-30B-A3B-Instruct & 30.3 & 54.3 \\
GLM-4.6V & 37.2 & 57.0 \\
MobileRL-9B & - & \textbf{80.2} \\
\midrule
Step-GUI-4B (Pass@1) & 31.9 & 63.9 \\
Step-GUI-8B (Pass@1) & 40.2 & 67.7 \\
\textbf{Step-GUI-4B (Pass@3)} & 40.4 & \underline{75.8} \\
\textbf{Step-GUI-8B (Pass@3)} & \underline{48.5} & \textbf{80.2} \\
\bottomrule
\end{tabular}
\vspace{-0.5em} 
\end{table*}

We evaluate our models on two challenging end-to-end environment benchmarks: OSWorld-Verified and AndroidWorld. Table~\ref{tab:online_env_benchmarks} presents a comprehensive comparison with state-of-the-art closed-source and open-source models.

\textbf{OSWorld-Verified.}~\cite{xie2024osworld} 
Due to the inherent instability of the OSWorld-Verified testing environment (including frequent VM crashes, substantial loading delays, and CAPTCHA interruptions), we employ the Pass@3 metric to mitigate failures caused by infrastructure issues rather than model limitations. 
On the OSWorld-Verified benchmark, our Step-GUI-8B achieves a score of 48.5, ranking second only to the closed-source Claude-4.5-sonnet (61.4) and significantly outperforming other competitive baselines. Notably, our model surpasses the powerful closed-source model OpenAI CUA o3 (23.0) by a substantial margin of +25.5 points, demonstrating superior reasoning and interaction capabilities in complex operating system environments. Our approach also outperforms specialized GUI understanding models such as SeedVL-1.5 (34.1), UI-TARS-1.5 (42.5), and UI-TARS-2 (47.5), as well as open-source alternatives like GUI-Owl-7B (34.9). Furthermore, compared to our baseline models Qwen3-VL-4B-Instruct (26.2) and Qwen3-VL-8B-Instruct (33.9), Step-GUI-8B achieves remarkable improvements of +22.3 and +14.6 points respectively, validating the effectiveness of our training pipeline and model design. The smaller Step-GUI-4B variant also demonstrates competitive performance with a score of 40.4, outperforming models with significantly larger parameters such as UI-TARS-1.5-7B (27.4).

\textbf{AndroidWorld.}~\cite{rawles2024androidworld} 
Given the significant stability challenges in the AndroidWorld testing environment (including unstable ADB communication, frequent Android emulator crashes, and system response delays), we employ the Pass@3 metric to exclude failures caused by infrastructure faults rather than model capability limitations. 
On the AndroidWorld benchmark, our Step-GUI-8B achieves state-of-the-art performance with a score of 80.2, tying with MobileRL-9B and establishing new performance standards for mobile GUI agents. This represents a substantial improvement over both closed-source and open-source methods. Specifically, our model outperforms UI-TARS-2 (73.3), Mobile-Agent-v3 (73.3), and the recent Gemini-2.5-Computer Use (69.7) by margins of +6.9, +6.9, and +10.5 points respectively. Compared to open-source GUI understanding models, Step-GUI-8B surpasses GUI-Owl-7B (66.4) by +13.8 points and SeedVL-1.5 (62.1) by +18.1 points. The performance gain over our baseline models is even more pronounced, with improvements of +34.9 points over Qwen3-VL-4B-Instruct (45.3) and +32.6 points over Qwen3-VL-8B-Instruct (47.6). Our smaller Step-GUI-4B variant achieves 75.8, securing the second-best performance among all tested models and outperforming all other open-source alternatives.

These results collectively demonstrate that our approach achieves exceptional performance across diverse GUI interaction scenarios, establishing new benchmarks for both desktop and mobile environments through our effective training pipeline and data construction methodology.

\begin{table*}[t]
\centering
\captionsetup{justification=justified, singlelinecheck=false}
\caption{Performance on AndroidDaily (Static) Benchmark. Best results are in \textbf{bold}, second best are \underline{underlined}.}
\label{tab:static_env_benchmarks}
\footnotesize
\setlength{\tabcolsep}{3pt} 
\resizebox{\textwidth}{!}{%
\begin{tabular}{l|ccccccccc}
\toprule
\textbf{Model} & CLICK & TYPE & SLIDE & AWAKE & INFO & COMPLETE & WAIT & LONG\_PRESS & AVG  \\
\midrule
GPT-4o & 12.65 & 52.41 & 31.17 & 10.84 & 0.00 & 44.85 & 5.88 & 0.00 & 17.73 \\
Claude-4.5-sonnet & 5.56 & 17.77 & \underline{50.65} & 1.33 & 0.00 & 53.48 & 21.18 & 0.00 & 10.90 \\
Gemini-2.5-Pro Thinking & 51.97 & 63.25 & 46.75 & 0.00 & 0.00 & 68.99 & 21.18 & \underline{33.33} & 43.74 \\
UI-TARS-1.5 & 84.43 & 73.49 & 48.05 & \underline{25.48} & 0.00 & 70.19 & 24.71 & \textbf{66.67} & 67.69 \\
\textbf{Step-GUI-4B} & \underline{87.11} & \underline{86.01} & 44.16 & \textbf{99.43} & \underline{52.01} & \underline{93.39} & \underline{74.12} & \textbf{66.67} & \underline{87.02} \\
\textbf{Step-GUI-8B} & \textbf{88.37} & \textbf{88.25} & \textbf{71.43} & \textbf{99.43} & \textbf{86.04} & \textbf{94.41} & \textbf{95.29} & \textbf{66.67} & \textbf{89.91} \\

\bottomrule
\end{tabular}
}
\vspace{-0.5em}
\end{table*}

\begin{table*}[t]
\centering
\captionsetup{justification=justified, singlelinecheck=false}
\caption{Performance on AndroidDaily (End-to-End) Benchmark.}
\label{tab:ad_end-to-end}

\resizebox{\textwidth}{!}{
\begin{tabular}{l
|ccc
|ccc
|ccc
|ccc
|c} 
\toprule
\multirow{2}{*}{\textbf{Model}} & 
\multicolumn{3}{c|}{\textbf{Task Type}} &
\multicolumn{3}{c|}{\textbf{Complexity}} &
\multicolumn{3}{c|}{\textbf{Ambiguity}} & 
\multirow{2}{*}{\textbf{Total}} \\ 

& Filter & Query & Analyze 
& Atomic & Comp. & Cond. 
& Low & Mid & High &  \\ 
\midrule

UI-TARS-1.5              & 57.64 & 65.97 & 36.71 
                         & 61.41 & 13.64 & 60.38 
                         & 57.05 & 54.90 & 57.89 & 56.64 \\

\textbf{Step-GUI-4B}   & 44.77 & 64.29 & 33.72 
                         & 54.03 & 19.61 & 42.86 
                         & 51.21 & 38.32 & 59.52 & 49.06  \\ 

\textbf{Step-GUI-8B}    & 52.50 & 63.82 & 32.95 
                         & 59.09 & 14.00 & 42.86  
                         & 54.08 & 44.55 & 61.54  & 52.50 \\ 
\bottomrule
\end{tabular}
}
\end{table*}

\noindent\textbf{Performance on AndroidDaily.}
We report evaluation results on both static action prediction and end-to-end task completion protocols.

\noindent\textit{Static Action Prediction.}
As shown in Table \ref{fig:android_daily_static}, Step-GUI-4B and Step-GUI-8B achieve state-of-the-art performance with average accuracies of 87.02\% and 89.91\% respectively, substantially outperforming all baseline models. UI-TARS-1.5 demonstrates competitive performance at 67.69\%, while general-purpose foundation models including GPT-4o (17.73\%), Claude-4.5-sonnet (10.90\%), and Gemini-2.5-Pro Thinking (43.74\%) lag significantly behind. This substantial performance gap indicates that Step-GUI and UI-TARS-1.5 possess superior domain-specific knowledge for Chinese mobile operations.

Across different action types, Step-GUI models consistently dominate. For basic interactions, both variants exceed 85\% accuracy on CLICK and TYPE actions. Notably, Step-GUI-8B achieves 71.43\% on SLIDE actions, significantly outperforming UI-TARS-1.5 (48.05\%), demonstrating better spatial reasoning for swipe gestures. For sophisticated operations including AWAKE, INFO, COMPLETE, WAIT, and LONG\_PRESS, Step-GUI models maintain strong performance with Step-GUI-8B achieving particularly high accuracy on INFO (86.04\%) and WAIT (95.29\%) actions.

\noindent\textit{End-to-End Task Completion.}
Table \ref{tab:ad_end-to-end} presents results where models interact with real applications. UI-TARS-1.5 achieves the highest success rate of 56.64\%, while Step-GUI-4B and Step-GUI-8B achieve comparable performance at 49.06\% and 52.50\% respectively. Despite UI-TARS-1.5's advantage in this setting, Step-GUI models demonstrate competitive end-to-end task completion capabilities, with Step-GUI-8B trailing by only 4.14 percentage points while significantly outperforming on static action prediction (89.91\% vs 67.69\%).

Analyzing by task type, all models perform best on Query tasks (requiring information retrieval) with success rates around 64-66\%, while Analyze tasks (demanding complex reasoning) prove most challenging with rates dropping to 33-37\%. For complexity dimensions, Atomic tasks show strong performance (54-61\%), while Composite tasks remain challenging for all models (14-20\%), indicating the difficulty of multi-step operation planning. Interestingly, examining instruction ambiguity reveals that Step-GUI-8B excels on high-ambiguity cases (61.54\%), surpassing UI-TARS-1.5 (57.89\%), demonstrating enhanced robustness in interpreting underspecified instructions, a critical capability for real-world deployment.

\begin{table*}[t]
\centering
\captionsetup{justification=justified, singlelinecheck=false}
\caption{Performance on Mainstream Multimodal Benchmarks. 
Best results are in \textbf{bold}, second best are \underline{underlined}.}
\label{tab:multimodal_benchmarks_adj_transposed}
\resizebox{\textwidth}{!}{
\begin{tabular}{l|ccccccc}
\toprule
\textbf{Benchmark} 
& \makecell{GLM-4.6V\\Flash}
& \makecell{Gemini-2.5\\flash-lite}
& \makecell{Qwen3-VL\\8B-Instruct}
& \makecell{Qwen3-VL-30B\\A3B-Instruct}
& \makecell{GPT5-nano\\minimal}
& \textbf{Step-GUI-4B} 
& \textbf{Step-GUI-8B} \\
\midrule

V*               &  -   &   -  & \underline{86.4} & \underline{86.4} & 66.5 & 79.1 & \textbf{89.0} \\
PhyX             &  -   &   -  & 28.3 & 31.1 & 24.0 & \underline{33.0} & \textbf{41.9} \\
OmniOCR          &  -   &   -  & \textbf{78.7} & 74.4 & 13.5 & 73.6 & \underline{77.1} \\
OCRBench         & 84.2 & 81.3 & \underline{89.7} & \textbf{90.8} & 70.1 & 84.6 & 88.0 \\
LogicVista       &  -   &   -  & 52.3 & \textbf{58.2} & 40.3 & 52.1 & \underline{53.7} \\
MMBench\_en      & 84.7 & 82.4 & \underline{89.9} & \textbf{90.8} & 51.6 & 87.2 & 89.1 \\
MMBench\_cn      & 85.8 &      & \underline{88.0} & \textbf{89.5} & 82.2 & 87.0 & \underline{88.0} \\
MMStar           & \textbf{72.9} & \underline{71.3} & 69.5 & 69.4 & 41.3 & 64.7 & 68.0 \\
MathVista\_{mini} & \underline{80.0} & 70.3 & 74.6 & \textbf{80.1} & 40.9 & 72.3 & 74.4 \\
CharXiv\_{DQ}     &  -   & 73.5 & 83.0 & \underline{85.5} & 64.4 & 84.1 & \textbf{86.3} \\
CharXiv\_{RQ}     &  -   & 44.6 & \underline{46.4} & \textbf{48.9} & 31.7 & 39.2 & 44.6 \\
SimpleVQA        &  -   & \underline{52.2} & 50.2 & \textbf{52.7} & 39.0 & 48.1 & 50.8 \\

\bottomrule
\end{tabular}}
\end{table*}

\noindent\textbf{Performance on General Multi-modal Benchmarks.}
To demonstrate that our model maintains strong general-purpose capabilities rather than being narrowly specialized for GUI tasks, we conduct comprehensive evaluations on mainstream multi-modal benchmarks. As shown in Table~\ref{tab:multimodal_benchmarks_adj_transposed}, \textbf{Step-GUI-8B} achieves superior or competitive performance across diverse evaluation scenarios compared to the base model Qwen3-VL-8B-Instruct and the larger Qwen3-30BA3B-Instruct.

Specifically, Step-GUI-8B achieves \textbf{89.0} on V*, surpassing both Qwen3-VL-8B-Instruct (86.4) and Qwen3-30BA3B-Instruct (86.4) by a significant margin. On OCRBench, our model scores \textbf{88.0}, outperforming Qwen3-VL-8B-Instruct (89.7) while maintaining comparable performance to Qwen3-30BA3B-Instruct (90.8). For mathematical reasoning evaluated on MathVista$_{mini}$, Step-GUI-8B achieves \textbf{74.4}, demonstrating strong reasoning capabilities that exceed the base model (74.6) while approaching the performance of the larger 30B model (80.1).
Notably, our model demonstrates consistent improvements across diverse benchmarks. Step-GUI-8B achieves \textbf{87.2} on MMBench\_en and \textbf{88.0} on MMBench\_cn for general multi-modal understanding, \textbf{68.0} on MMStar for visual reasoning, and strong performance on document understanding tasks (CharXiv$_{DQ}$: \textbf{86.3}, OmniOCR: \textbf{77.1}), indicating that GUI-oriented training does not compromise general multi-modal understanding capabilities. Furthermore, Step-GUI-4B, despite having only half the parameters, maintains competitive performance across most benchmarks, achieving \textbf{84.6} on OCRBench, which validates the effectiveness of our training approach.

These results demonstrate that our training methodology successfully enhances GUI-specific capabilities while \textit{preserving and even improving} general-purpose multi-modal understanding abilities. This characteristic is crucial for practical deployment, as it enables the model to handle diverse real-world tasks beyond GUI interaction without requiring separate specialized models.

\subsection{Detailed Analysis}

\subsubsection{Self-Evolving Training Dynamics}
\begin{figure}
    \centering
    \includegraphics[width=1.0\linewidth]{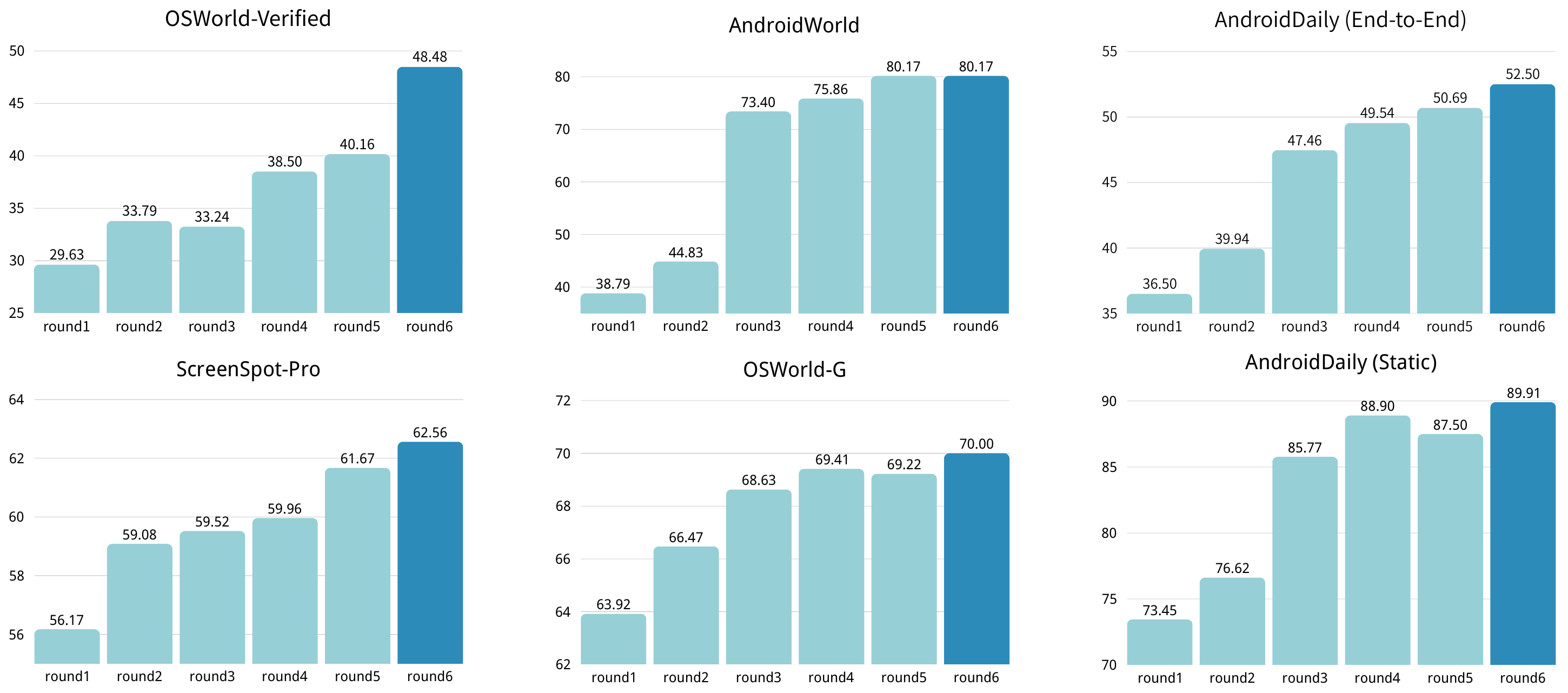}
    \caption{Performance Evolution of Step-GUI-8B Across Six Self-Evolving Training Rounds.}
    \label{fig:self_evolving}
\end{figure}

Based on the experimental data spanning Rounds 1 through 6 (as illustrated in the figure\ref{fig:self_evolving}), we validate the effectiveness of the self-evolving training framework proposed in Step-GUI. 

\noindent\textbf{"Phase Transition" on AndroidWorld.}
On the AndroidWorld benchmark, we observe a remarkable performance jump from Round 2 (44.83\%) to Round 3 (73.40\%). We attribute this surge to the Date Flow 1 (Generation) mechanism within the CSRS. 
During the initial phase (Round 1), the model addresses fundamental knowledge gaps via the Cold-Start stage. 
As the model's capabilities reach a critical threshold in Round 2, the CSRS begins to effectively capture and verify successful long-horizon trajectories generated during exploration. 
These self-discovered, high-quality trajectories enriched with Chain-of-Thought serve as high-value signals that flow back into the training pipeline, catalyzing an explosive growth in performance.

\noindent\textbf{Steady Advancement on OSWorld.}
 Compared to the mobile domain, desktop tasks on OSWorld present greater complexity. 
 Our data indicates a steady, quasi-linear improvement on OSWorld performance, rising from 29.63\% to 46.26\%. 
 This trajectory highlights the efficacy of the Date Flow 1 (Refinement). 
 By employing "Rejection Sampling" and "Self-Distillation" on challenging samples, the system continuously identifies model weaknesses near complex decision boundaries.
 It subsequently converts failed trajectories into Knowledge Data for targeted reinforcement, thereby enabling sustained breakthroughs in long-horizon, complex tasks.

Furthermore, as illustrated by the AndroidDaily (End-to-End) results in figure~\ref{fig:self_evolving}, the performance curve exhibits a smooth upward trend (49.06\% $\rightarrow$ 65.06\%). 
This stability stems from the CSRS design, which discards traditional step-level rewards in favor of rewards assigned via objective verification of final task success. 
This sparse yet high-confidence signal ensures the purity of the data fed back into the system, thereby guaranteeing robustness across multiple training iterations.
In conclusion, the synchronous capability improvements observed across diverse domains indicate that Step-GUI's self-evolving system successfully establishes a robust, general-purpose multimodal foundation through closed-loop training iterations. 
The system effectively transmutes computational resources into data quality, which subsequently evolves into model intelligence. 
This process marks a paradigm shift, transitioning the model from a reliance on human priors to a mechanism driven by self-exploration and reflection.

\begin{figure}
    \centering
    \includegraphics[width=1.0\linewidth]{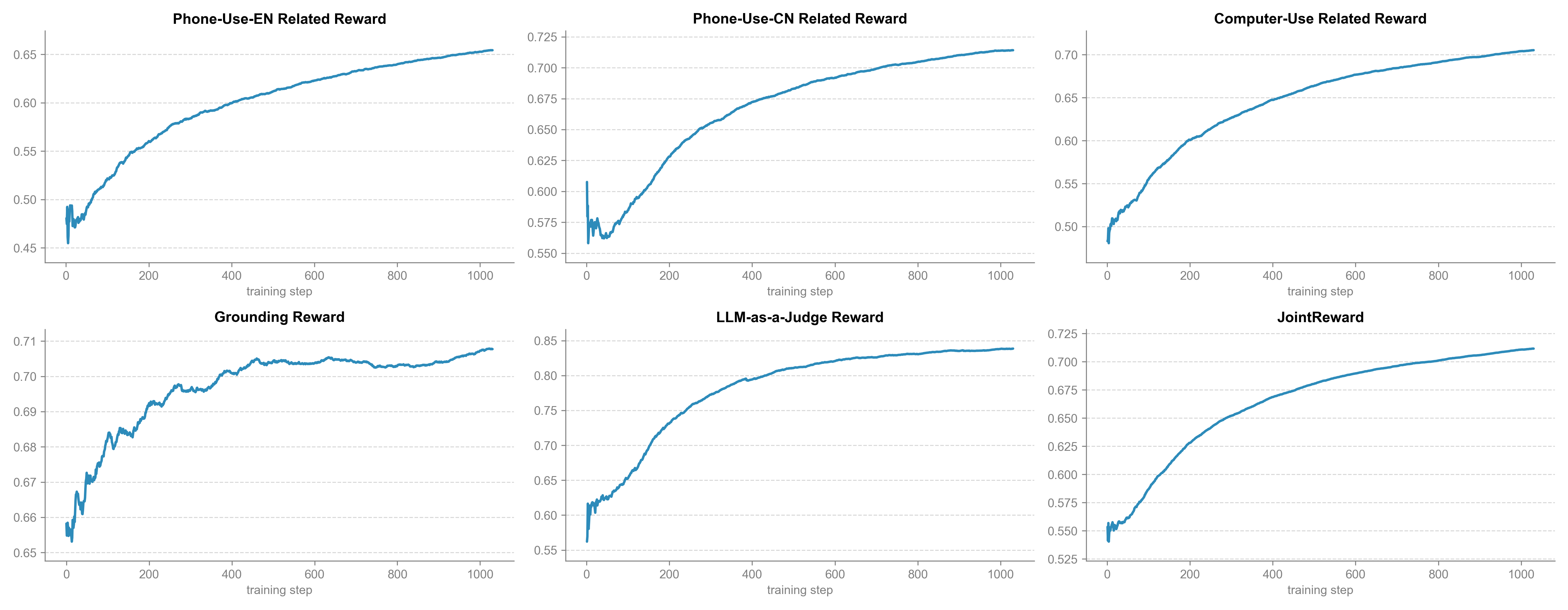}
    \captionsetup{justification=justified, singlelinecheck=false}
    \caption{Reward Dynamics during RLVR Training. The smooth, monotonic ascent across task-specific sub-rewards (Top) and the aggregate JointReward (Bottom Right) demonstrates stable convergence without oscillatory collapse. The strong correlation between LLM-as-a-Judge Reward and objective metrics confirms the policy improves genuine capabilities while avoiding reward hacking.}
    \label{fig:joint_reward}
\end{figure}

\subsubsection{RLVR Training: Stability and Convergence}

To rigorously assess the stability and convergence properties of the Step-GUI during RLVR stage, we monitor the optimization dynamics through a comprehensive suite of reward signals and off-policy correction metrics. 
These metrics quantify the distributional shift between the rollout policy ($\pi_{\text{rollout}}$) and the training policy ($\pi_{\text{train}}$), validating the robustness of our training.

\noindent\textbf{Reward Consistency and Alignment.}
As illustrated in Figure~\ref{fig:joint_reward}, the training process exhibits a remarkably smooth and monotonic ascent across all signal sources. The JointReward rises steadily without the oscillatory behavior or collapse often associated with RL training. 
Notably, the sub-rewards for specific capabilities, including phone-use-en, phone-use-cn, computer-use, and grounding, show high correlation with the LLM-as-a-Judge reward curve, which is demonstrated to align well with human annotations.
This synchronization confirms that our composite reward function effectively anchors the model's optimization trajectory to human-aligned reasoning, preventing reward hacking.

\begin{figure}
    \centering
    \includegraphics[width=1.0\linewidth]{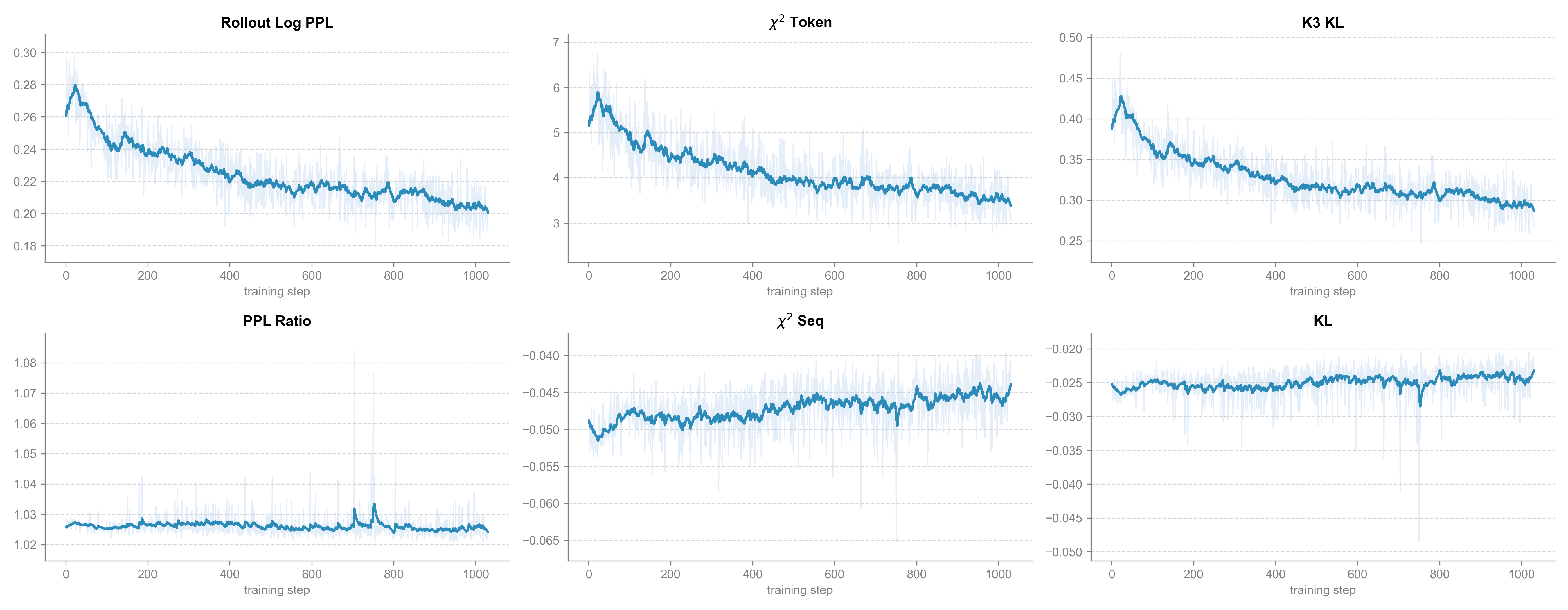}
    \captionsetup{justification=justified, singlelinecheck=false}
    \caption{Off-Policy Correction and Stability Diagnostics. The downward trends in Rollout Log PPL and $\chi^2$ (Token/Seq) variances indicate increasing model confidence and reduced gradient estimation noise. Complementing this, the decreasing K3-KL divergence demonstrates effective suppression of extreme policy deviations, while the low and stable PPL Ratio and KL confirm high synchronization between training and rollout policies, ensuring updates remain within a safe trust region.}
    \label{fig:joint_others}
\end{figure}

\noindent\textbf{Policy Synchronization.}
We scrutinize the off-policy divergence through the diagnostic metrics presented in Figure~\ref{fig:joint_others}.
First, the Rollout Log PPL displays a consistent decline (from $\sim 0.28$ to $\sim 0.20$), indicating that the rollout policy progressively gains confidence and reduces perplexity in its generated trajectories.
Concurrently, the PPL Ratio, quantifying the deviation between $\pi_{\text{train}}$ and $\pi_{\text{rollout}}$, remains consistently low and stable, fluctuating strictly between $1.02$ and $1.03$.
This dual observation confirms that even as the model becomes more deterministic, the training policy maintains high synchronization with the rollout policy, ensuring that gradient updates remain valid and effectively "on-policy" despite the inherent lag in data collection.

\noindent\textbf{Tail Risk Suppression.}
To measure distributional divergence, we monitor both standard KL divergence and the K3-KL estimator ($E[e^{\Delta} - \Delta - 1]$), which is highly sensitive to extreme deviations in the policy's tail distribution. 
While standard KL remains stable, the K3-KL metric in Figure~\ref{fig:joint_others} exhibits a distinct downward trend (from $\sim 0.45$ to $\sim 0.30$). 
This demonstrates that the model progressively minimizes "surprise" or extreme deviations from the exploration trajectory, keeping updates within a safe trust region.

\noindent\textbf{Variance Reduction.}
We further employ Importance Sampling (IS) with a sequence-truncate strategy to stabilize gradients.
The effectiveness of this mechanism is evidenced by both the token-level and sequence-level $\chi^2$ metrics, which approximate the variance of importance weights $Var(\rho)$.
As shown in Figure~\ref{fig:joint_others}, the $\chi^2$ Token metric exhibits a consistent decline from $\sim 5.5$ to $\sim 3.5$, indicating that local weight fluctuations are progressively smoothed out.
Complementing this, the $\chi^2$ Sequence maintains a stable, low-magnitude profile throughout training.
This joint behavior implies that the importance weights are becoming more uniform at both the fine-grained token level and the holistic sequence level.
Consequently, the reduction in gradient estimation noise directly substantiates the smooth convergence observed in the reward curves.

\begin{figure}
    \centering
    \includegraphics[width=1.0\linewidth]{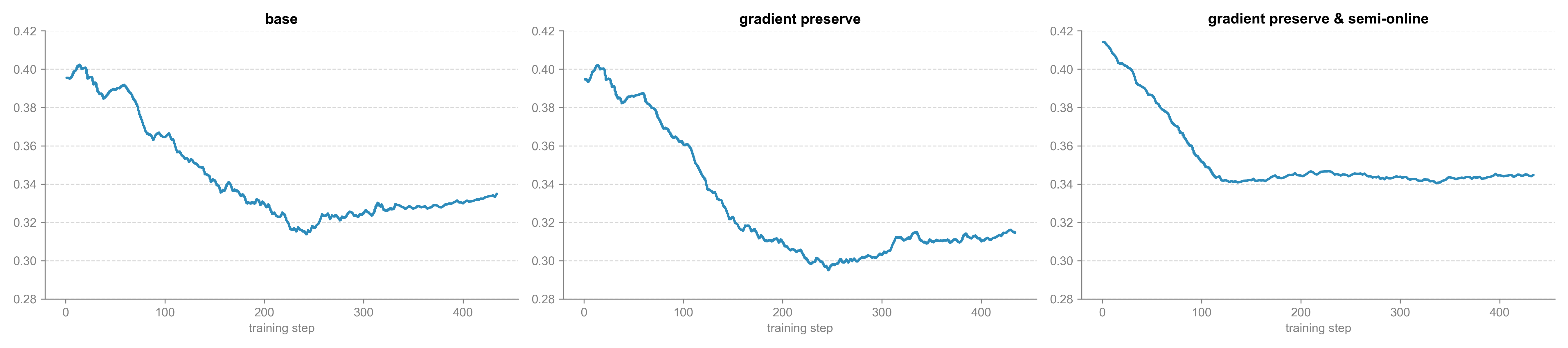}
    \captionsetup{justification=justified, singlelinecheck=false}
    \caption{Evolution of Policy Entropy under Different Training Strategies. Baseline GRPO (Left) shows rapid entropy decay followed by stabilization. Gradient Preservation (Middle) further reduces entropy, indicating tighter exploitation but risking mode collapse. Semi-Online strategy (Right) significantly revitalizes entropy by injecting ground-truth hints into failed rollouts, effectively disrupting narrow distributions and sustaining healthy exploration as a counter-force to premature convergence.}
    \label{fig:entropy}
\end{figure}
\noindent\textbf{Training Entropy.}
To evaluate exploration-exploitation dynamics, we analyze policy entropy evolution in Figure~\ref{fig:entropy}. 
The baseline GRPO (Left) exhibits rapid decay followed by stabilization. 
Integrating Gradient Preservation (Middle) leads to a further reduction in entropy, indicating tighter exploitation of clipped samples but a heightened risk of mode collapse. 
In contrast, the Semi-Online strategy (Right) significantly revitalizes entropy levels. 
By injecting ground-truth hints into failed rollouts, this mechanism effectively disrupts narrow distributions and sustains healthy exploration, serving as a crucial counter-force to premature convergence.

\section{Related Work}
\subsection{Reasoning with Language Models}

Chain-of-thought (CoT) prompting~\cite{wei2022chain_cot} demonstrates that eliciting intermediate reasoning steps significantly enhances large language models' performance on complex reasoning tasks. Building on this foundation, synthetic prompting~\cite{shao2023synthetic_synprompt} and STaR~\cite{zelikman2022star} explore automated generation of reasoning demonstrations: the former employs backward-forward construction to create high-quality exemplars, while the latter bootstraps rationale datasets from minimal seed examples and introduces rationalization, training models to justify correct answers post-hoc. Tree of Thoughts~\cite{yao2023tree_tot} extends CoT to breadth-first search over alternative reasoning branches, enabling backtracking and global solution evaluation. Quiet-STaR~\cite{zelikman2024quiet_quietstar} further generalizes this paradigm to learn reasoning from unstructured text across all token positions through parallel sampling.

To enhance reasoning reliability, recent work introduces verification mechanisms. Self-consistency~\cite{wang2022self_consistency} samples multiple reasoning paths and selects the most frequent answer, while self-verification~\cite{weng2023large_self_verification} and Self-Refine~\cite{madaan2023self_selfrefine} implement iterative refinement through self-critique. Addressing hallucinations in knowledge-intensive scenarios, CoT-RAG~\cite{li2025cot_rag} proposes knowledge graph-driven CoT generation with expert-provided decision trees that encapsulate domain reasoning logic, while self-refinement-enhanced knowledge retrieval~\cite{niu2024mitigating} identifies high-hallucination tokens via attribution analysis and corrects inaccuracies through targeted retrieval. DDCoT~\cite{zheng2023ddcot} extends this to multimodal reasoning by partitioning responsibilities between vision models and language models through duty-distinct prompting.
For self-improvement through data generation, impossible distillation~\cite{jung2023impossibledistill} filters low-quality samples generated from suboptimal models, while ReST~\cite{gulcehre2023reinforced_rest} employs offline reinforcement learning to iteratively improve policies from self-generated data. Despite these advances, efficiently obtaining high-quality trajectory and reasoning data for target domains remains challenging; we address this through a CSRS-centered data flywheel that ensures annotation quality while maintaining scalability. 

\subsection{GUI Agents}

Recent work, including UI-TARS~\cite{qin2025ui_uitars1}, UI-TARS-2~\cite{wang2025ui_uitars2}, OpenCUA~\cite{wang2025opencua}, GUI-Owl~\cite{ye2025mobile_mobileagentv3}, and UITron~\cite{zeng2025uitron}, systematically investigates data curation, cleaning pipelines, and training methodologies spanning pretraining and post-training phases. Concurrently, mainstream foundation models such as Claude series~\cite{claudeopus45}, Seed series~\cite{guo2025seed1}, and Qwen series~\cite{yang2025qwen3} begin natively integrating GUI interaction capabilities, underscoring the strategic importance of this research direction. Most GUI agents follow the ReAct paradigm~\cite{yao2022react}, observe, reason via CoT, then act, with recent systems extending this through multi-agent collaboration and hierarchical architectures. Notable examples include Mobile-Agent-v3~\cite{ye2025mobile_mobileagentv3}, which introduces specialized agents (Manager, Worker, Reflector, Notetaker) for knowledge evolution and reflective reasoning, and other systems~\cite{liu2025pc,song2025coact1,zhang2025ufo2,agashe2025agents2} that explore hierarchical planning, code-visual hybrid control, and compositional agent coordination. We propose the Step-GUI series models, achieving state-of-the-art performance among models of similar size, while Step-GUI-4B can run fully locally on consumer-grade hardware, enabling deployment in privacy-sensitive scenarios. 

To evaluate these advances, the community develops both static benchmarks including ScreenSpot~\cite{cheng2024seeclick}, ScreenSpot-v2~\cite{wu2024atlas}, ScreenSpot-Pro~\cite{li2025screenspotpro}, FuncPred~\cite{li2025autogui}, MoTIF~\cite{burns2022dataset_motif}, RefExp~\cite{rastogi2021uibert_refexp}, LlamaTouch~\cite{zhang2024llamatouch}, VWB-AG~\cite{liu2024visualwebbench_vwb}, and VWB-EG~\cite{liu2024visualwebbench_vwb}, that primarily assess element grounding and task planning accuracy through prediction-annotation matching, and interactive benchmarks such as AndroidWorld~\cite{rawles2024androidworld} and OSWorld~\cite{xie2024osworld} that enable agents to execute tasks within realistic or virtualized operating system environments with multi-dimensional task completion evaluation. In contrast to these efforts, we focus on scalable reasoning trajectory synthesis and efficient agent architecture design for real-world GUI automation scenarios.
\section{Conclusion}
In this work, we present a holistic framework advancing practical GUI agents across three dimensions: data, deployment, and evaluation. 
We introduce the Calibrated Step Reward System (CSRS), a self-evolving pipeline that drastically reduces annotation costs while enabling the training of SOTA Step-GUI models (4B/8B).
To standardize deployment, we propose GUI-MCP, a hierarchical privacy-centric protocol that balances execution efficiency with on-device data security. 
Finally, we contribute AndroidDaily, a benchmark grounded in authentic usage patterns to rigorously evaluate real-world utility.
Together, these innovations bridge the critical gap between research capabilities and reliable daily assistance.

\newpage
\section*{Contributors}

Our contributor list is primarily sorted alphabetically by first name, with the project leader placed at the very end. $^\star$ denoting GELab team, and $\dagger$ indicates the project leader. 

\noindent{\large\textbf{Core Contributors}}

\noindent{\textbf{Algorithm:}}
Haolong Yan$^\star$, Jia Wang$^\star$, Xin Huang$^\star$, Yeqing Shen$^\star$, Ziyang Meng$^\star$, Zhimin Fan, Kaijun Tan$^\star$$^\dagger$

\noindent{\textbf{Infra:}}
Jin Gao, Lieyu Shi, Mi Yang, Shiliang Yang, Zhirui Wang, Brian Li$^\dagger$, Kang An$^\dagger$

\noindent{\textbf{Application:}}
Chenyang Li, Lei Lei, Mengmeng Duan, Danxun Liang$^\dagger$

\noindent{\textbf{Data:}}
Guodong Liu, Hang Cheng, Hao Wu, Jie Dong, Junhao Huang, Mei Chen, Renjie Yu, Shunshan Li, Xu Zhou, Yiting Dai, Yineng Deng, Yingdan Liang, Zelin Chen, Wen Sun$^\dagger$

\noindent{\large\textbf{Contributors}}

Chengxu Yan, Chunqin Xu, Dong Li, Fengqiong Xiao, Guanghao Fan, Guopeng Li, Guozhen Peng, Hongbing Li, Hang Li, Hongming Chen, Jingjing Xie, Jianyong Li, Jingyang Zhang, Jiaju Ren, Jiayu Yuan, Jianpeng Yin, Kai Cao, Liang Zhao, Liguo Tan, Liying Shi, Mengqiang Ren, Min Xu, Manjiao Liu, Mao Luo, Mingxin Wan, Na Wang, Nan Wu, Ning Wang, Peiyao Ma, Qingzhou Zhang, Qiao Wang, Qinlin Zeng, Qiong Gao, Qiongyao Li, Shangwu Zhong, Shuli Gao, Shaofan Liu, Shisi Gao, Shuang Luo, Xingbin Liu, Xiaojia Liu, Xiaojie Hou, Xin Liu, Xuanti Feng, Xuedan Cai, Xuan Wen, Xianwei Zhu, Xin Liang, Xin Liu, Xin Zhou, Yifan Sui, Yingxiu Zhao, Yixun Zhang, Yukang Shi, Yunfang Xu, Yuqing Zeng, Zejia Weng, Zhonghao Yan, Zhiguo Huang, Zhuoyu Wang, Zihan Yan

\noindent{\large\textbf{Supervisor:}}
Zheng Ge, Jing Li, Yibo Zhu, Binxing Jiao, Xiangyu Zhang, Daxin Jiang

\bibliography{reference}
\newpage
\appendix
\section*{Appendix}

\begin{figure}[h]
    \centering
    \includegraphics[width=\linewidth]{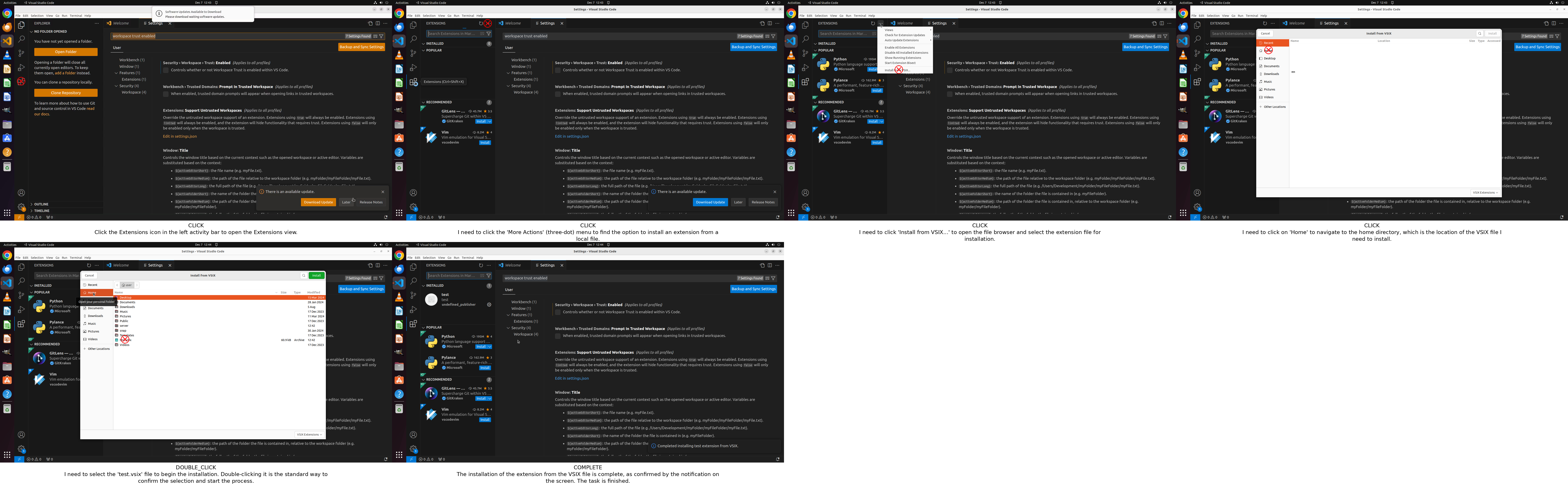}
    \captionsetup{justification=justified, singlelinecheck=false}
    \caption{OSWorld Trajectory 1: Please help me install an extension in VS Code from a local VSIX file "/home/user/test.vsix".}
    \label{fig:osworld_trajectory1}
\end{figure}

\begin{figure}[h]
    \centering
    \includegraphics[width=\linewidth]{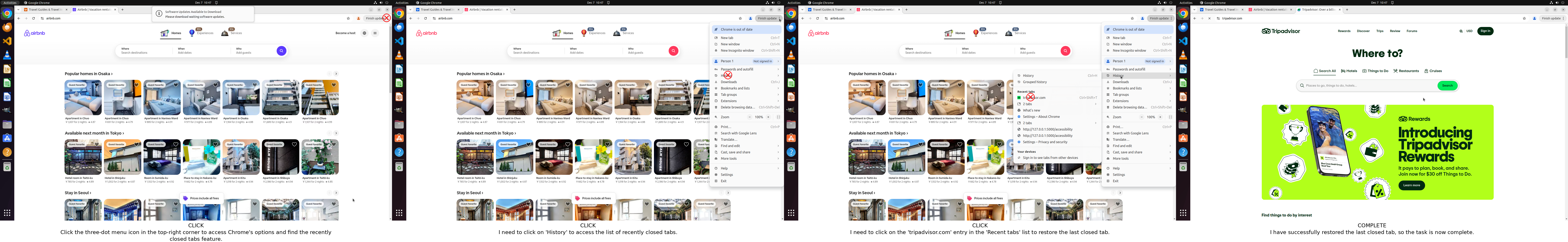}
    \captionsetup{justification=justified, singlelinecheck=false}
    \caption{OSWorld Trajectory 2: Can you make my computer bring back the last tab I shut down?}
    \label{fig:osworld_trajectory2}
\end{figure}

\begin{figure}[h]
    \centering
    \includegraphics[width=\linewidth]{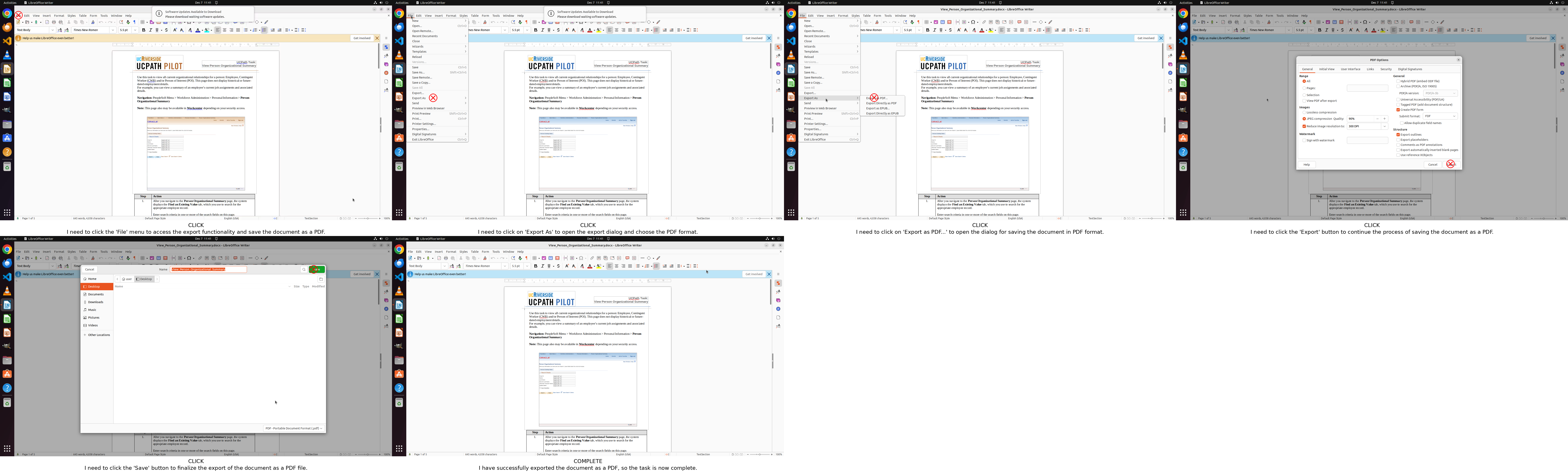}
    \captionsetup{justification=justified, singlelinecheck=false}
    \caption{OSWorld Trajectory 3: Export the current document into PDF, keep the file name}
    \label{fig:osworld_trajectory3}
\end{figure}

\begin{figure}[h]
    \centering
    \includegraphics[width=\linewidth]{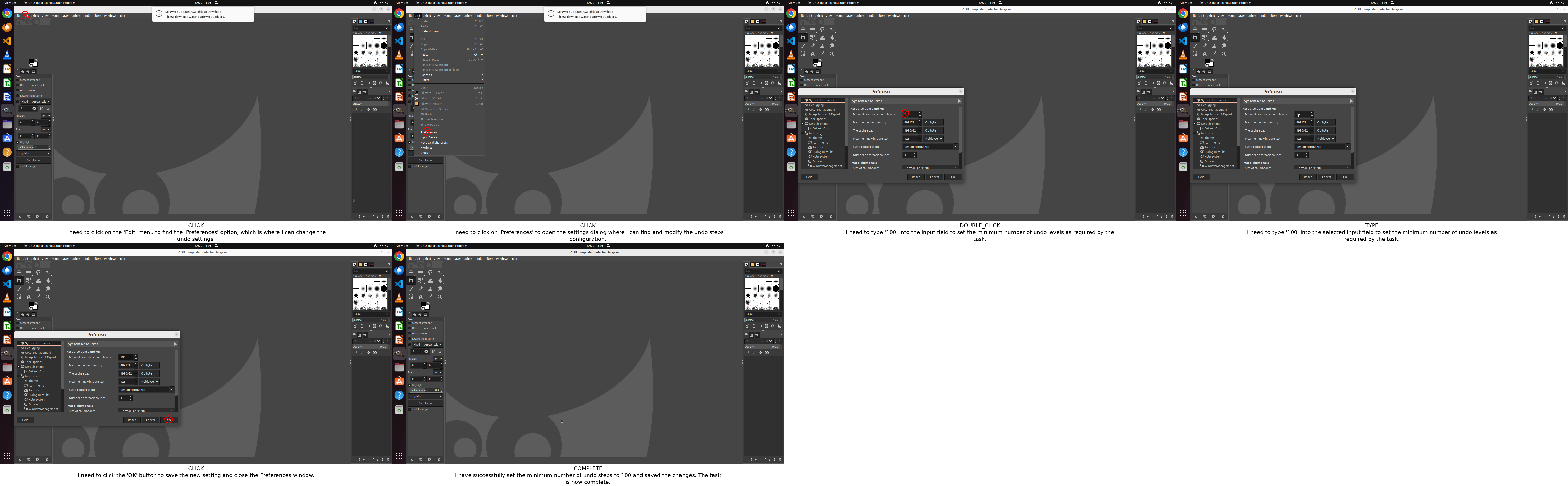}
    \captionsetup{justification=justified, singlelinecheck=false}
    \caption{OSWorld Trajectory 4: Set the minimum number of undo steps to 100.}
    \label{fig:osworld_trajectory4}
\end{figure}

\begin{figure}[h]
    \centering
    \includegraphics[width=\linewidth]{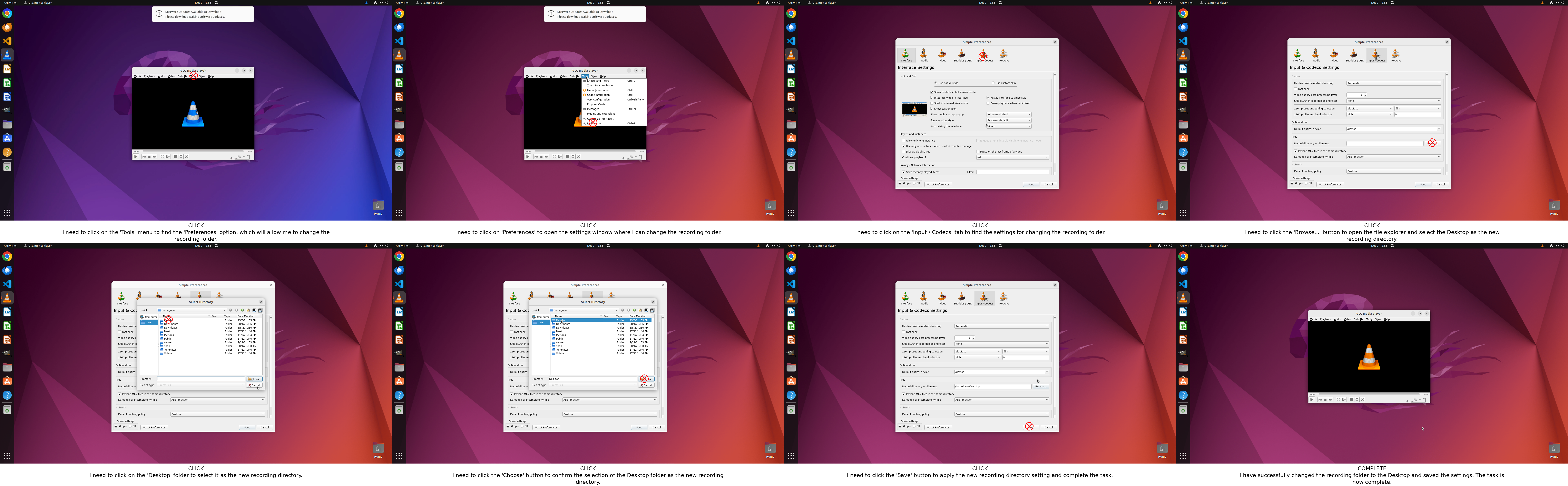}
    \captionsetup{justification=justified, singlelinecheck=false}
    \caption{OSWorld Trajectory 5: Help me modify the folder used to store my recordings to Desktop}
    \label{fig:osworld_trajectory5}
\end{figure}

\begin{figure}[h]
    \centering
    \includegraphics[width=\linewidth]{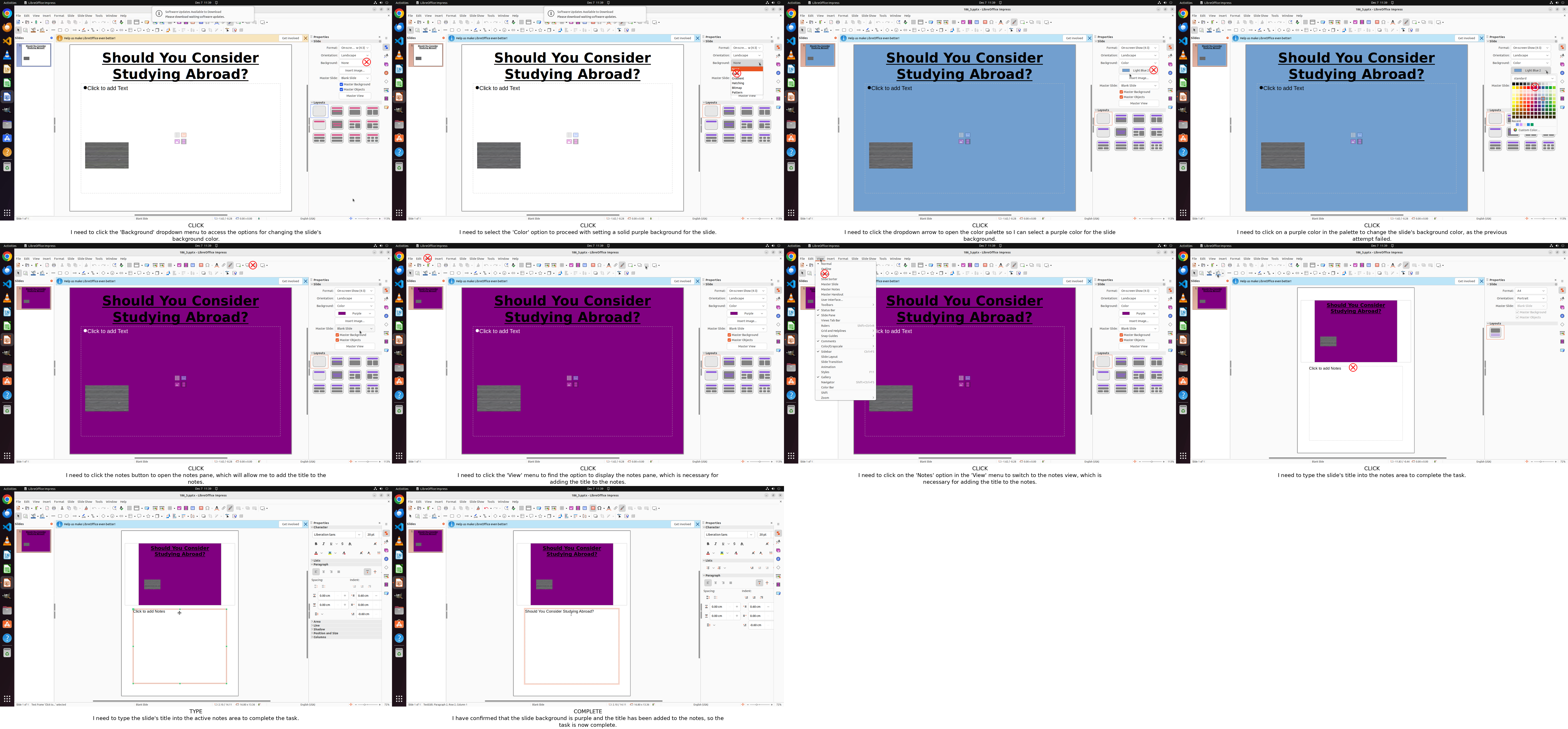}
    \captionsetup{justification=justified, singlelinecheck=false}
    \caption{OSWorld Trajectory 6: Give the slide a purple background color. Add the title to note.}
    \label{fig:osworld_trajectory6}
\end{figure}

\begin{figure}[h]
    \centering
    \includegraphics[width=0.9\linewidth]{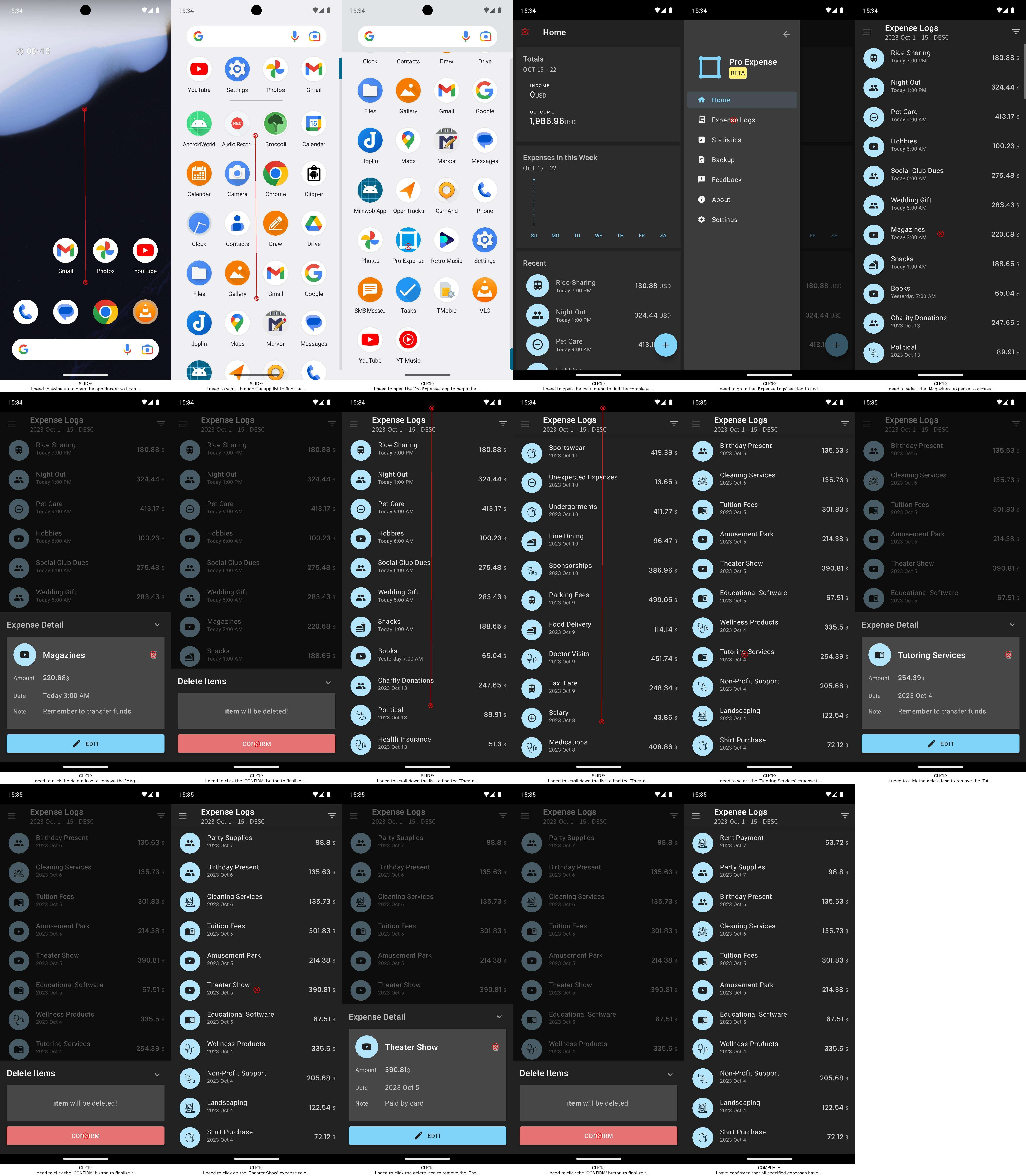}
    \captionsetup{justification=justified, singlelinecheck=false}
    \caption{AndroidWorld Trajectory 1: Delete the following expenses from pro expense: Tutoring Services, Magazines, Theater Show.}
    \label{fig:androidworld_trajectory1}
\end{figure}

\begin{figure}[h]
    \centering
    \includegraphics[width=\linewidth]{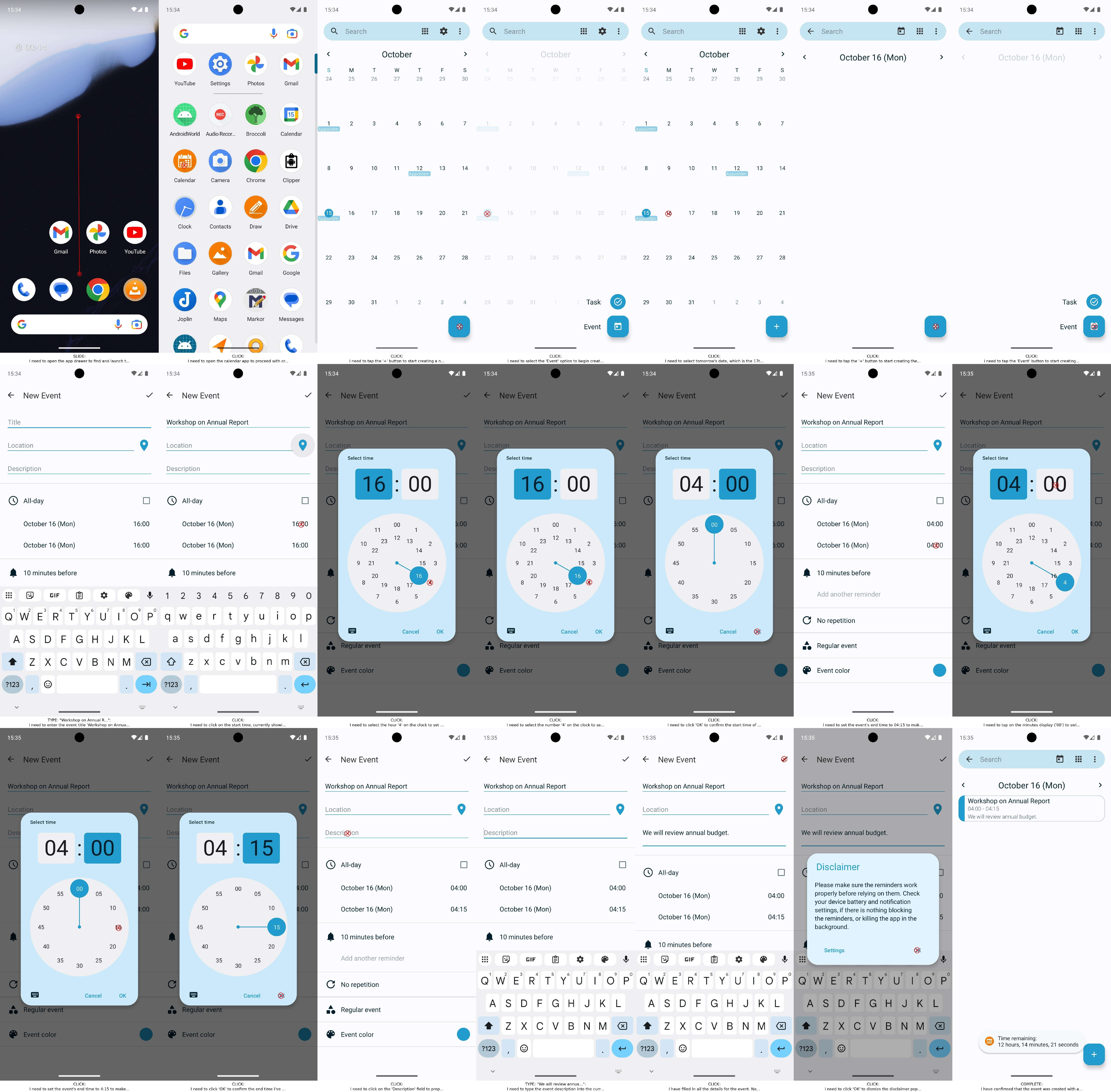}
    \captionsetup{justification=justified, singlelinecheck=false}
    \caption{AndroidWorld Trajectory 2: In Simple Calendar Pro, create a calendar event for tomorrow at 4h with the title 'Workshop on Annual Report' and the description 'We will review annual budget.'. The event should last for 15 mins.}
    \label{fig:androidworld_trajectory2}
\end{figure}

\begin{figure}[h]
    \centering
    \includegraphics[width=\linewidth]{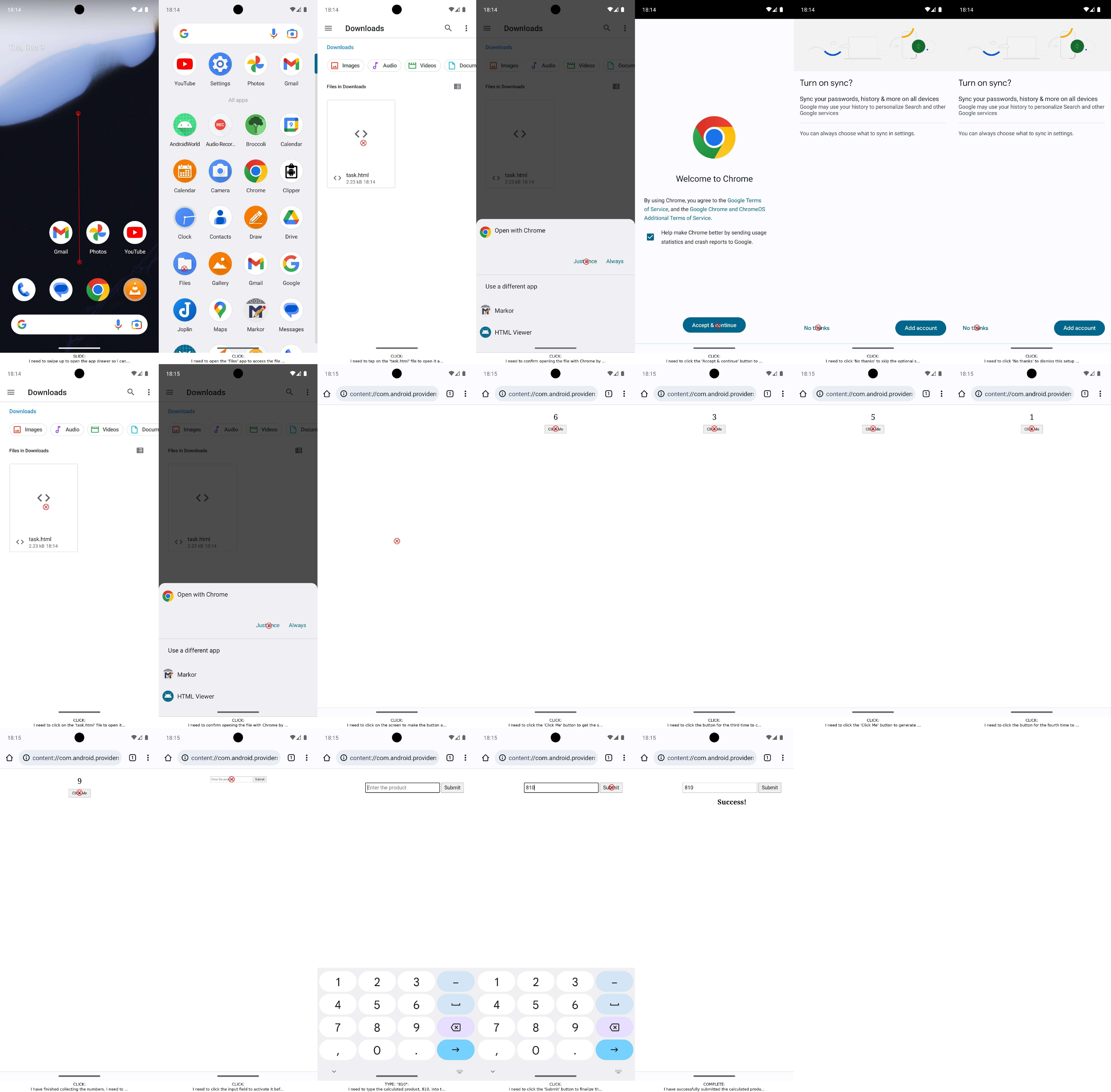}
    \captionsetup{justification=justified, singlelinecheck=false}
    \caption{AndroidWorld Trajectory 3: Open the file task.html in Downloads in the file manager; when prompted open it with Chrome. Then click the button 5 times, remember the numbers displayed, and enter their product in the form.}
    \label{fig:androidworld_trajectory3}
\end{figure}

\begin{figure}[h]
    \centering
    \includegraphics[width=0.7\linewidth]{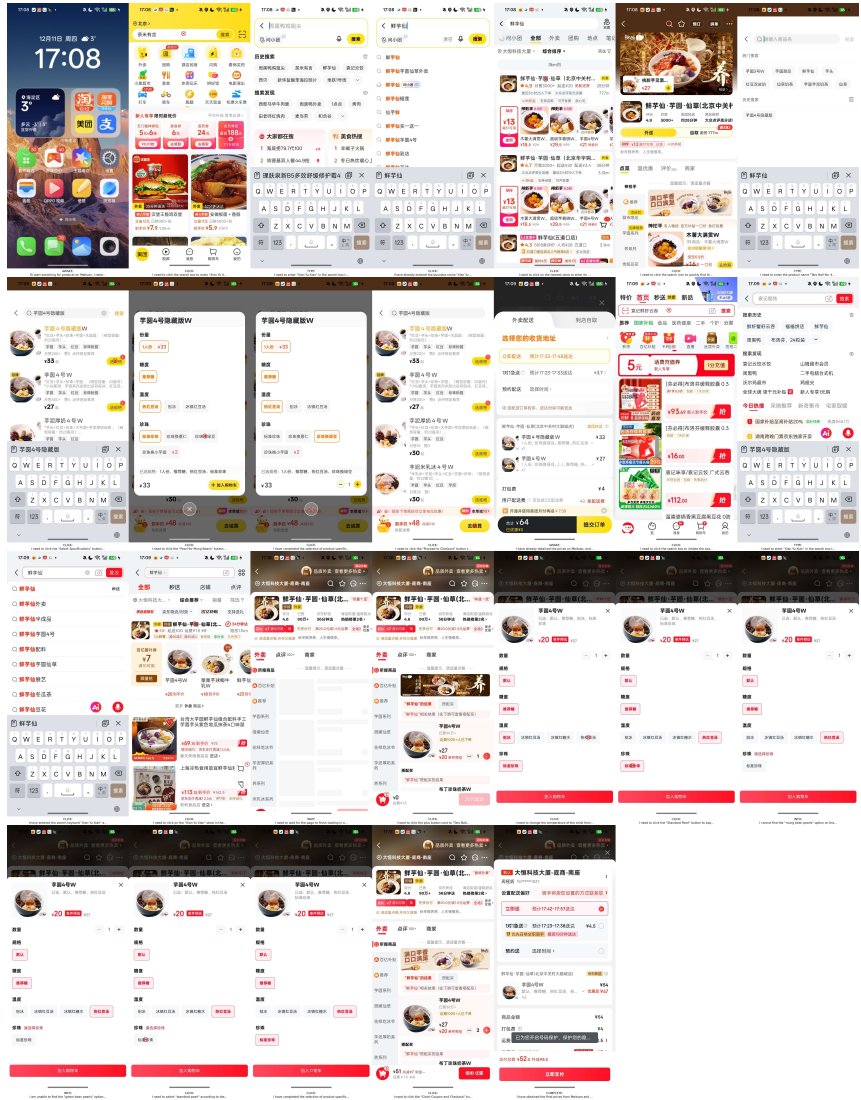}
    \captionsetup{justification=justified, singlelinecheck=false}
    \caption{AndroidDaily (End-to-End) Trajectory 1: Check both Meituan Delivery and JD FlashEx. Buy the hidden version of Taro Balls No. 4 at the Xianyu Xian store closest to Beijing Daheng Technology Building. For the specifications, use hot red bean soup as the base and replace the pearl boba with mung beans. Which platform offers a better price?}
    \label{fig:androiddaily_trajectory1}
\end{figure}

\begin{figure}[h]
    \centering
    \includegraphics[width=\linewidth]{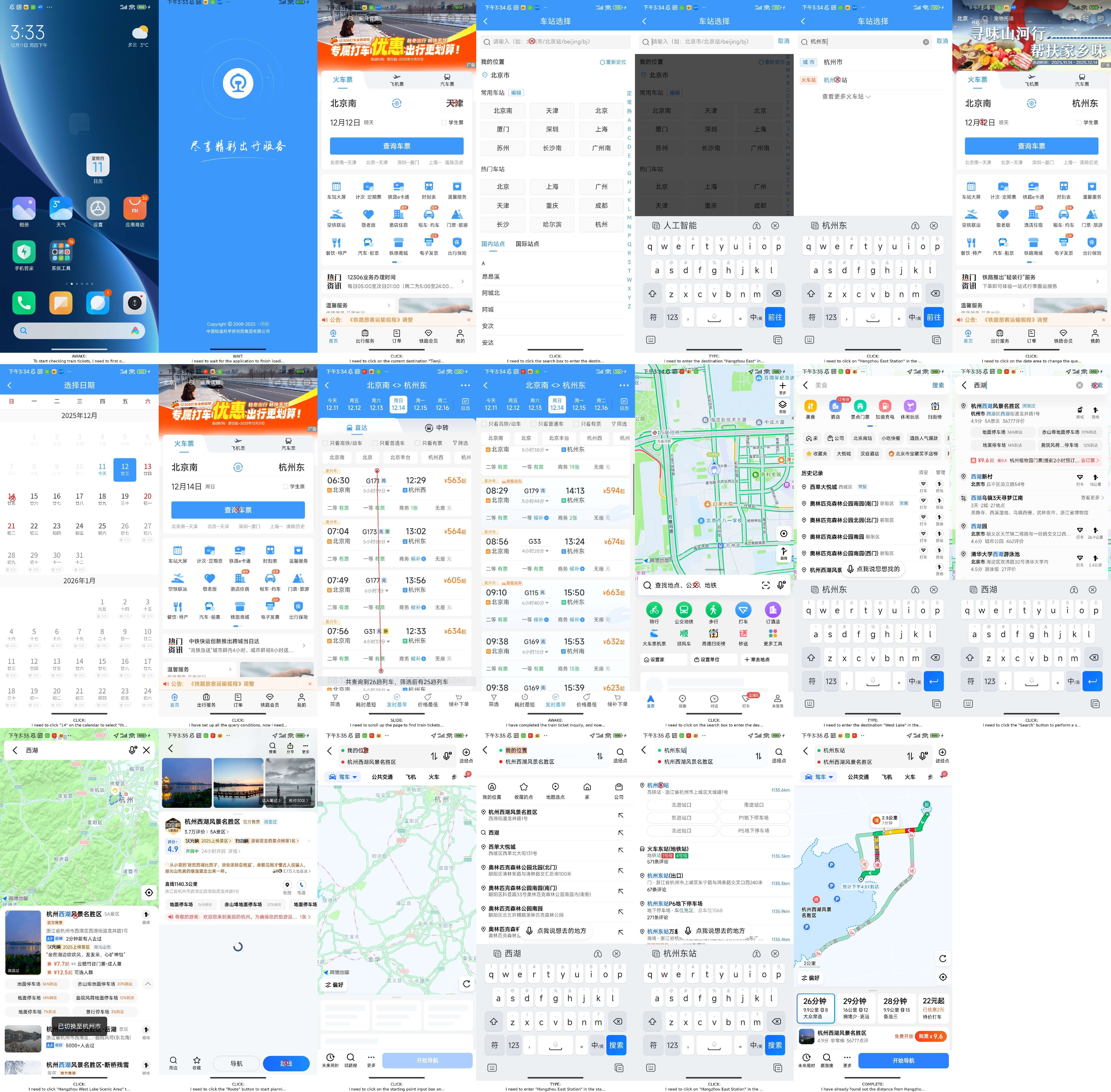}
    \captionsetup{justification=justified, singlelinecheck=false}
    \caption{AndroidDaily (End-to-End) Trajectory 2: First, check the high-speed trains arriving at Hangzhou East Railway Station around 9 a.m. the day after tomorrow via Railway 12306. Then, use Amap to find out the distance from Hangzhou East Railway Station to West Lake.}
    \label{fig:androiddaily_trajectory2}
\end{figure}

\begin{figure}[h]
    \centering
    \includegraphics[width=\linewidth]{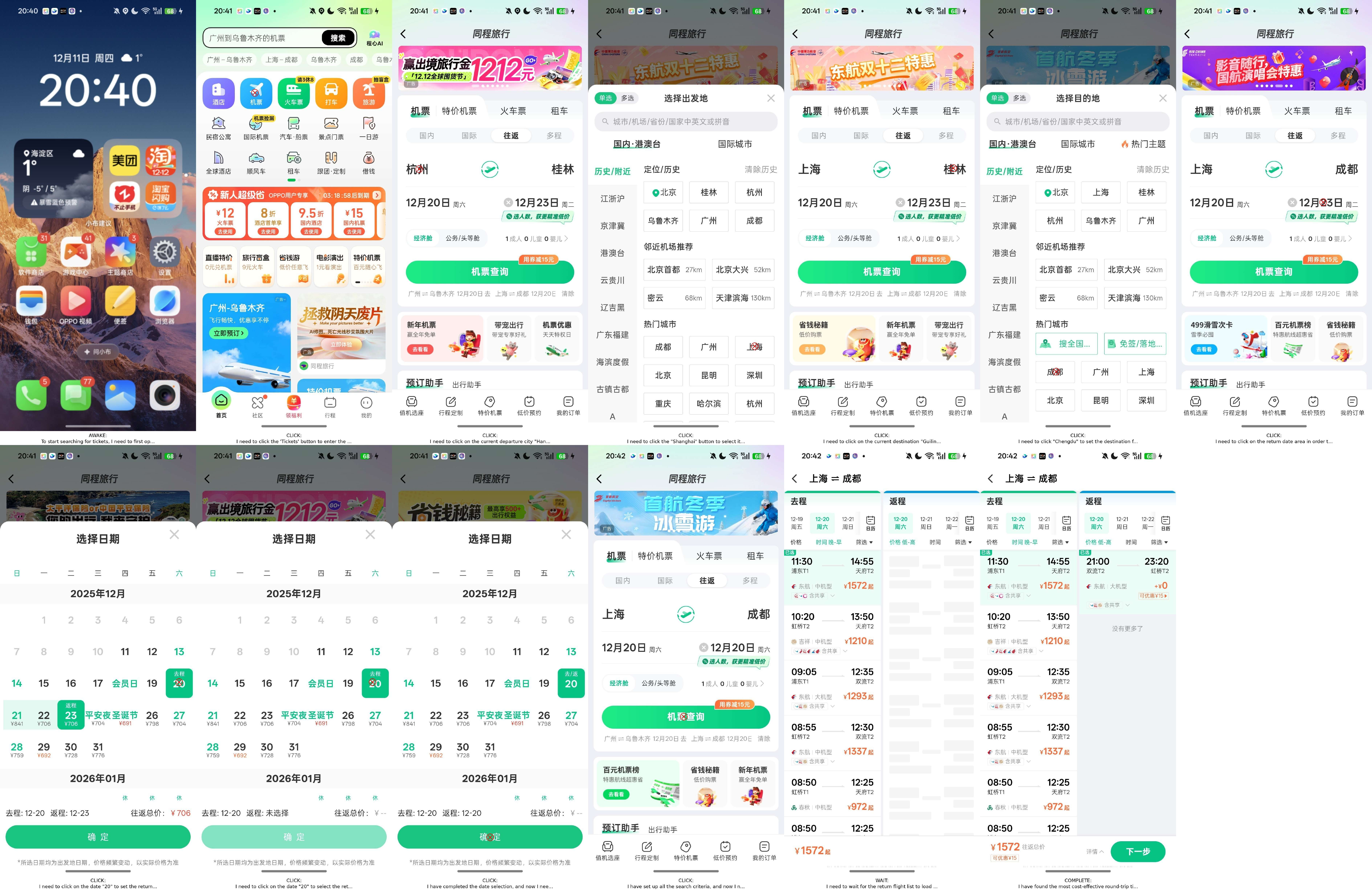}
    \captionsetup{justification=justified, singlelinecheck=false}
    \caption{AndroidDaily (End-to-End) Trajectory 3: Look for round-trip flights departing from Shanghai to Chengdu on December 20th via Tongcheng Travel and see which airline offers the best deal.}
    \label{fig:androiddaily_trajectory3}
\end{figure}

\begin{figure}[h]
    \centering
    \includegraphics[width=\linewidth]{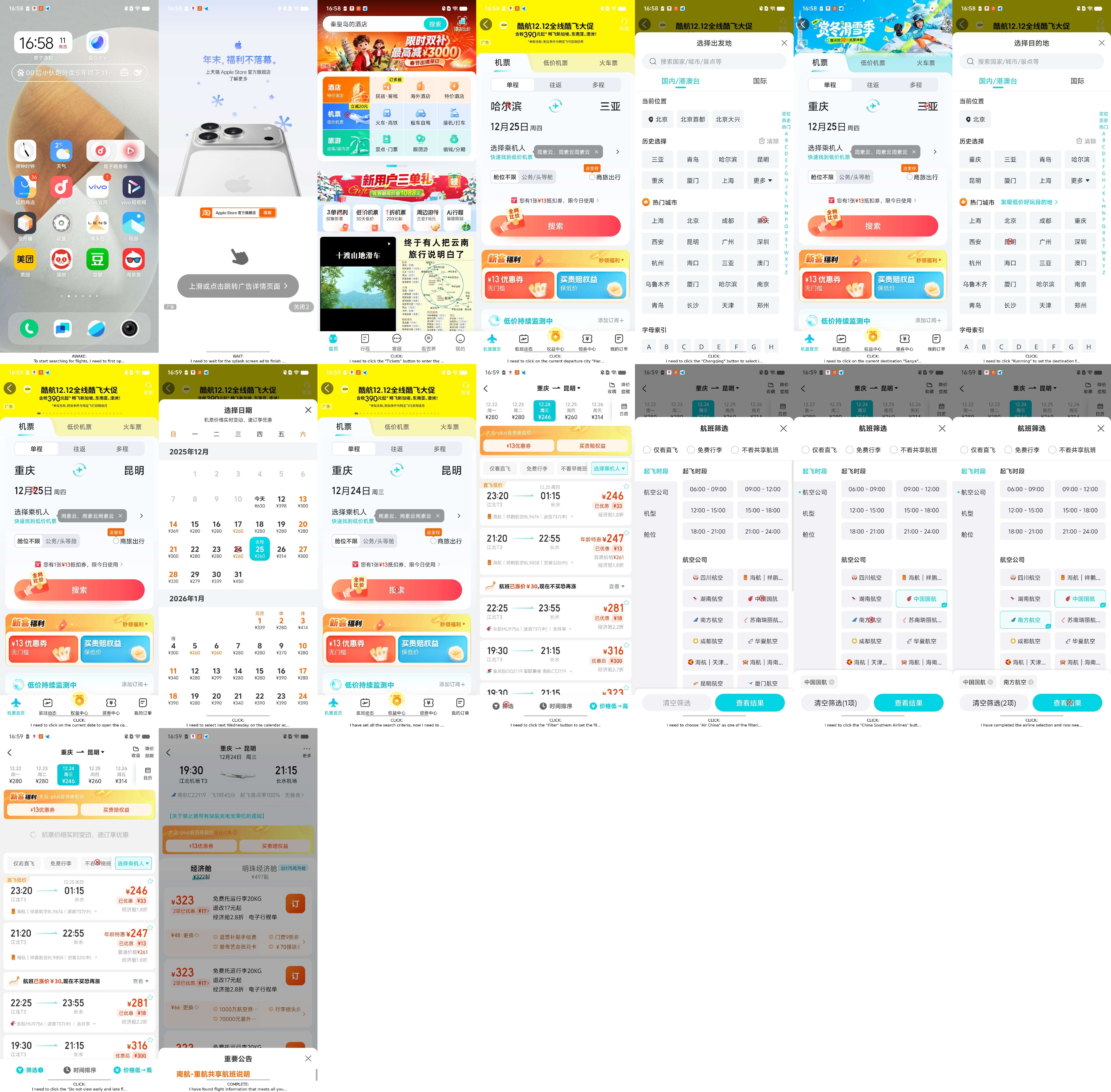}
    \captionsetup{justification=justified, singlelinecheck=false}
    \caption{AndroidDaily (End-to-End) Trajectory 4: Look for flights from Chongqing to Kunming on Qunar that depart next Wednesday. Choose either China Southern Airlines or Air China, and exclude red-eye flights.}
    \label{fig:androiddaily_trajectory4}
\end{figure}

To provide concrete evidence of our system's capabilities in real-world scenarios, we present representative successful task trajectories from the OSWorld benchmark (Figure~\ref{fig:osworld_trajectory1} - Figure~\ref{fig:osworld_trajectory6}), AndroidWorld benchmark (Figure~\ref{fig:androidworld_trajectory1} - Figure~\ref{fig:androidworld_trajectory3}) and AndroidDaily (End-to-End) (Figure~\ref{fig:androiddaily_trajectory1} - Figure~\ref{fig:androiddaily_trajectory4}). These trajectories demonstrate the agent's ability to accurately perceive screen elements, reason about task requirements, and execute appropriate actions in the correct sequence. Each trajectory shows the visual state of the GUI along with the corresponding agent actions throughout the task execution, highlighting the effectiveness of our proposed approach in understanding and interacting with diverse graphical user interfaces.

\subsection*{A. Trajectories on OSWorld}

\begin{figure}[h]
    \centering
    \includegraphics[width=\linewidth]{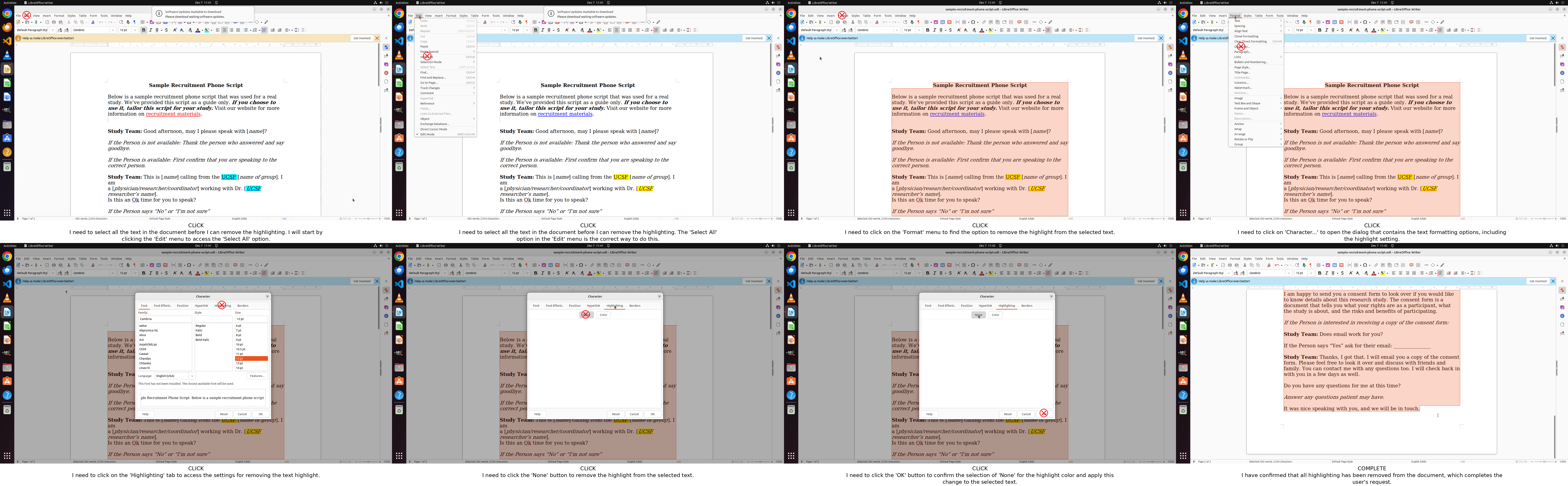}
    \captionsetup{justification=justified, singlelinecheck=false}
    \caption{OSWorld Trajectory 7: I have been editing my document and some words that needed to be rewritten are highlighted in yellow. As I fixed those words, please help me remove all highlight. I want to make sure that there is no highlight word.}
    \label{fig:osworld_trajectory7}
\end{figure}

\begin{figure}[h]
    \centering
    \includegraphics[width=\linewidth]{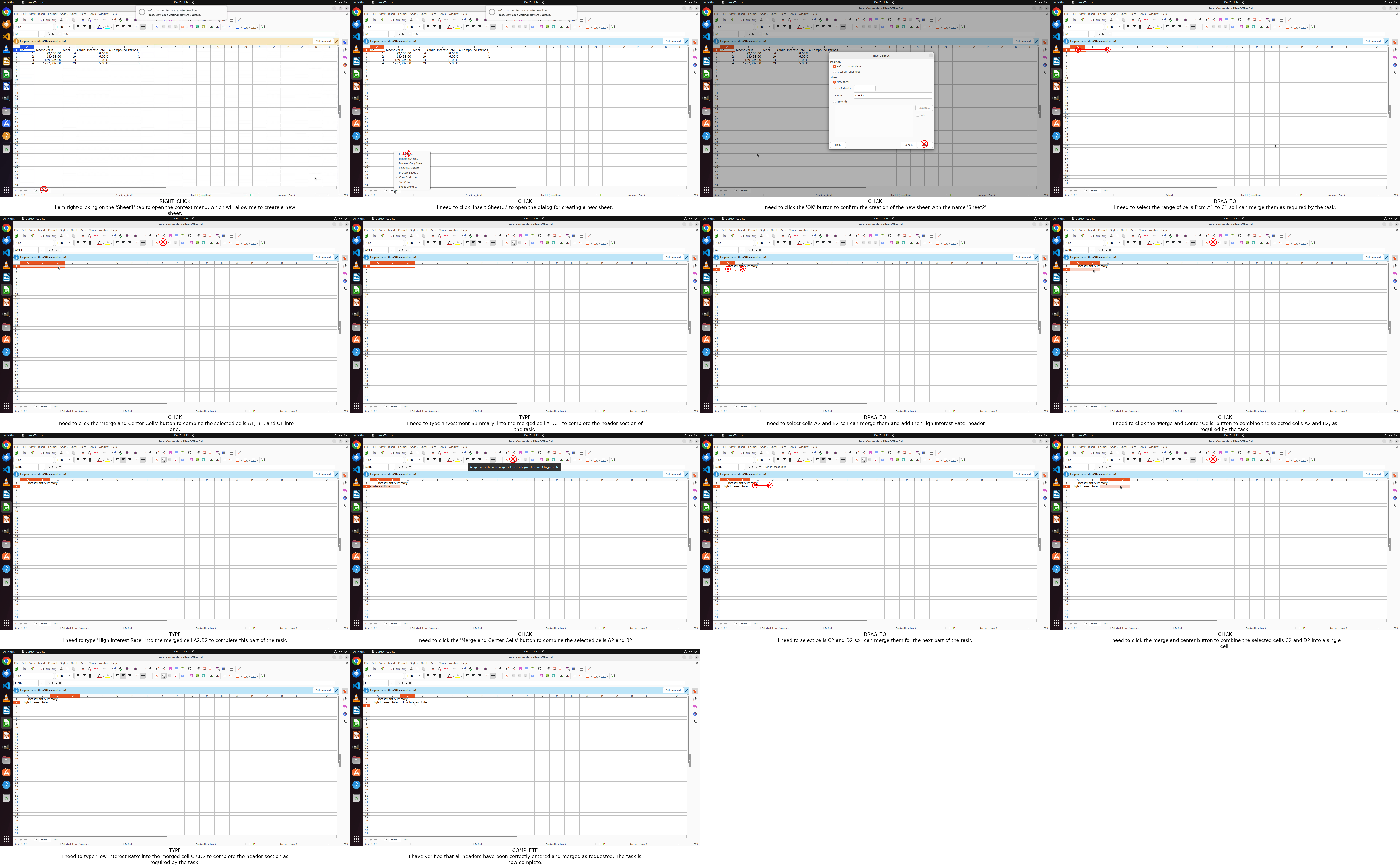}
    \captionsetup{justification=justified, singlelinecheck=false}
    \caption{OSWorld Trajectory 8: Create a new sheet named "Sheet2" and merge cells A1:C1 to write the header "Investment Summary". Beneath that, merge cells A2:B2 to write "High Interest Rate" and merge cells C2:D2 to form "Low Interest Rate".}
    \label{fig:osworld_trajectory8}
\end{figure}

\clearpage
\subsection*{B. Trajectories on AndroidWorld}

\clearpage
\subsection*{C. Trajectories on AndroidDaily (End-to-End) }

\clearpage

\subsection*{D. GUI-MCP Example}
\begin{figure}[h]
    \centering
    \includegraphics[
    width=\linewidth,
    trim=10 10 10 10,
     clip
    ]
    {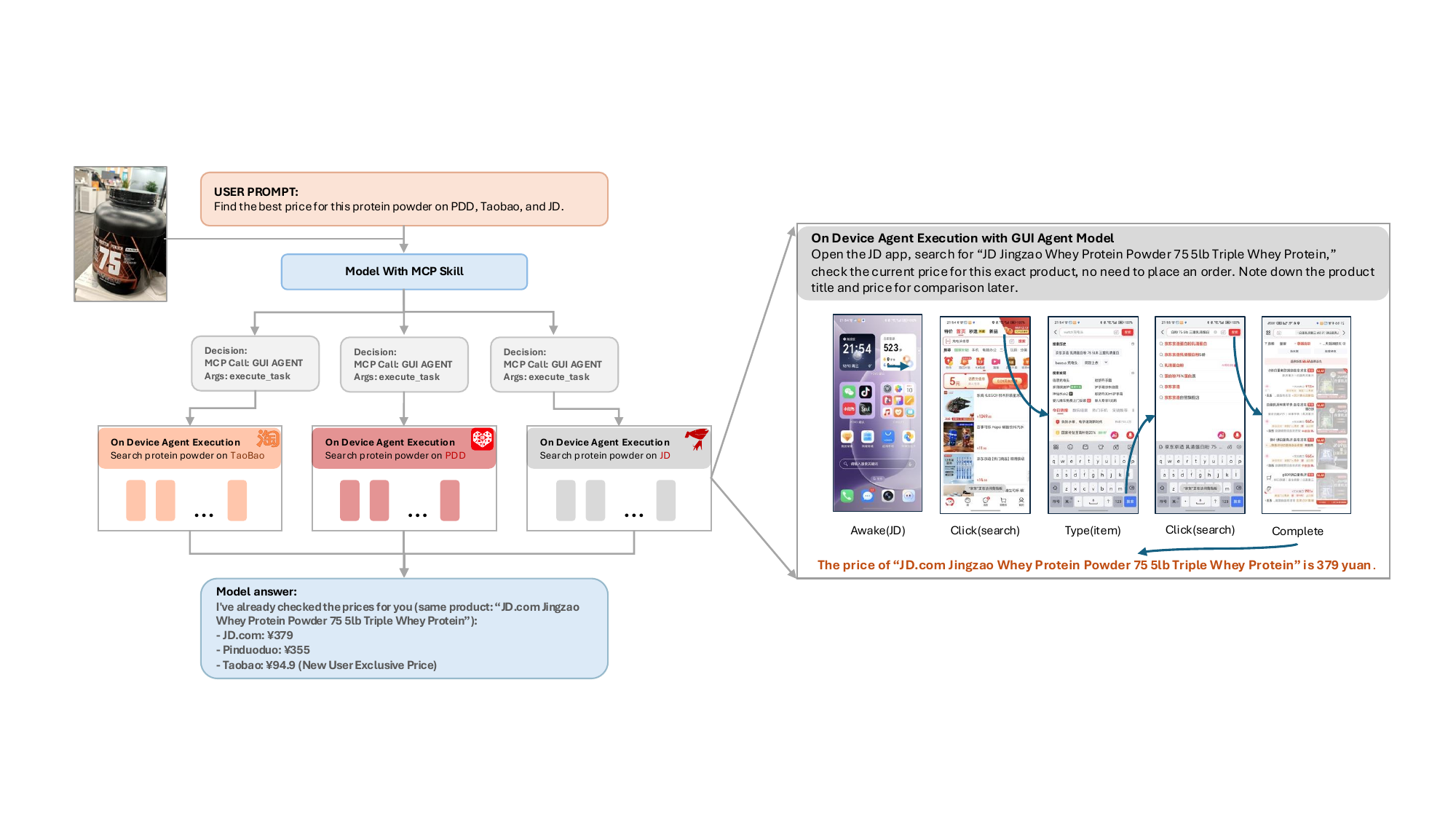}
    \captionsetup{justification=justified, singlelinecheck=false}
    \caption{GUI-MCP enables LLMs to perform complex cross-platform mobile tasks. The framework decomposes a price comparison query into parallel platform-specific search tasks, executes them via on-device GUI agents, and synthesizes results into a structured response.}
    \label{fig:placeholder}
\end{figure}

To demonstrate the practical utility of GUI-MCP, we present a cross-platform price comparison task that showcases the framework's parallel task decomposition, unified multi-platform control, and autonomous on-device execution capabilities. As illustrated in Figure~\ref{fig:placeholder}, given a user query to find the best price for a protein powder product across three major Chinese e-commerce platforms (PDD, Taobao, and JD.com), the main LLM leverages MCP's standardized interface to decompose this high-level request into three independent platform-specific search subtasks and invokes the \texttt{execute\_task} interface in parallel, significantly reducing overall task completion time.

The right panel details the execution trace of the JD.com agent, which autonomously completes a five-step interaction sequence: launching the application, activating the search interface, inputting the query "JD Jingzao Whey Protein Powder 7.5lb Triple Whey Protein", submitting the search, and extracting the price (¥379). This entire execution flow is performed by the locally deployed GUI specialist model without requiring additional interaction with the external main LLM, thereby reducing API costs and latency while preserving user privacy. Upon completion of all parallel tasks, the system synthesizes results from all three platforms—JD.com (¥379), Pinduoduo (¥355), and Taobao (¥94.9)—into a coherent response, demonstrating GUI-MCP's capability to seamlessly orchestrate complex cross-platform operations.

\end{document}